\documentclass{article} 
\usepackage[final]{colm2025_conference}

\usepackage{microtype}
\usepackage{hyperref}
\usepackage{url}
\usepackage{booktabs}

\usepackage{lineno}

\usepackage{graphicx}
\usepackage{amsmath}
\usepackage{subcaption}

\definecolor{darkblue}{rgb}{0, 0, 0.5}
\hypersetup{colorlinks=true, citecolor=darkblue, linkcolor=darkblue, urlcolor=darkblue}

\title{Can a Crow Hatch a Falcon?\\Lineage Matters in Predicting Large Language Model Performance}

\author{Takuya Tamura, Taro Yano, Masafumi Enomoto \& Masafumi Oyamada \\
NEC Corporation\\
\texttt{\{tamura-takuya, taro\_yano, masafumi-enomoto, oyamada\}@nec.com} \\
}

\begin{document}

\ifcolmsubmission
\linenumbers
\fi

\maketitle


\begin{abstract}
Accurately forecasting the performance of Large Language Models (LLMs) before extensive fine-tuning or merging can substantially reduce both computational expense and development time. Although prior approaches like scaling laws account for global factors such as parameter size or training tokens, they often overlook explicit \emph{lineage} relationships—i.e., which models are derived or merged from which parents. In this work, we propose a novel \textbf{Lineage-Regularized Matrix Factorization (LRMF)} framework that encodes ancestral ties among LLMs via a graph Laplacian regularizer. By leveraging multi-hop parent--child connections, LRMF consistently outperforms conventional matrix factorization and collaborative filtering methods in both instance-level and benchmark-level performance prediction. Our large-scale study includes \textbf{2,934 publicly available Hugging Face models} and \textbf{21,000+ instances} across \textbf{6 major benchmarks}, 
showing that the introduction of lineage constraints yields up to \textbf{0.15–0.30 higher Pearson correlation} coefficients with actual performance compared to baseline methods.
Moreover, LRMF effectively addresses the \emph{cold-start problem}, providing accurate estimates for newly derived or merged models even with minimal data. This lineage-guided strategy thus offers a resource-efficient way to inform hyperparameter tuning, data selection, and model combination in modern LLM development.
\end{abstract}

\section{Introduction}
\label{sec:intro}

Building and refining a Large Language Model (LLM) is an increasingly expensive venture. Developers face critical decisions such as model size, the amount and type of training data, and how to allocate parameters for fine-tuning. Moreover, \emph{model merging} (or model ``souping'') is emerging as a promising strategy: distinct models or checkpoints are combined at the parameter level to create new variants with complementary strengths \citep{pmlr-v162-wortsman22a-model-soup}. Each of these steps---training, fine-tuning, and merging---often demands large-scale exploration and incurs a high computational cost.

If one could \emph{accurately predict} the performance of a proposed LLM variant---even before running large-scale optimization---it would profoundly reduce the search space. Such foresight would streamline decisions about (1) the right model size and data volume for pre-training, (2) fine-tuning hyperparameters and data composition (e.g., domain mix), and (3) whether combining certain model checkpoints is worthwhile. Indeed, prior work has sought to provide partial solutions. Classic \emph{scaling laws} \citep{kaplan2020scalinglawsneurallanguage,hoffmann2022an} predict an LLM's loss on the training corpus as a function of parameter size and token count. These laws help developers extrapolate the compute needed to reach a desired perplexity or training loss. Subsequent studies extend these ideas to specific benchmarks, e.g., MMLU~\citep{hendrycks2021measuring_mmlu,wang2024mmlupro}, BBH~\citep{suzgun-etal-2023-challenging}, or MATH~\citep{hendrycks2021measuring_MATH}, by fitting regression models or matrix factorization frameworks to empirical performance data \citep{zhang-etal-2024-collaborative,polo2024tinybenchmarks,kipnis2025metabench}.

However, real-world LLMs rarely arise in isolation. In practice, a single \emph{base model} often spawns multiple variants via continual pre-training, domain-specific fine-tuning, preference tuning (e.g., RLHF), or parameter-level merging of multiple checkpoints. Such derivation or \emph{lineage} relationships typically imply that ``child'' models inherit certain performance characteristics from their parents. Despite this, previous performance-prediction work has largely ignored the explicit structure of these relationships, focusing instead on coarse factors such as model size or training tokens. As a result, the potential advantage of modeling multi-hop ancestry---i.e., ``who was the parent's parent?''---remains underexplored.

In this paper, we propose a novel way to incorporate such lineage information into performance prediction. Specifically, we introduce a \textbf{Lineage-Regularized Matrix Factorization (LRMF)} framework that treats the known derivation paths among models as a \emph{lineage graph}. Within a standard matrix factorization procedure (where models form the rows and benchmarks/tasks form the columns), we add a graph Laplacian regularization term that explicitly constrains models with direct or multi-hop parent--child ties to lie ``close'' in the latent space. The resulting method can more accurately propagate performance signals from well-evaluated parent models to their lesser-known descendants.

We conduct a large-scale empirical study encompassing \textbf{2,934 publicly available Hugging Face models} and \textbf{21,000+ instances} across \textbf{6 major benchmarks}. We demonstrate that incorporating lineage constraints---i.e., explicitly modeling which models were fine-tuned or merged from which parents---consistently boosts predictive accuracy. In fact, compared to standard matrix factorization baselines, 
our method improves the Pearson correlation coefficient with true performance by up to \textbf{0.15–0.30} on certain tasks,
affirming that lineage is a key factor in understanding and forecasting LLM behaviors.

Moreover, by leveraging explicit lineage relationships (parent--child relations or merges), our method LRMF tackles the \textbf{cold-start problem}, enabling accurate performance estimates for newly derived or merged models without costly full re-benchmarking. Indeed, our experiments reveal that even with as few as \textbf{10 labeled instances per model}, we can achieve near state-of-the-art Pearson correlation with fully evaluated baselines---a remarkable improvement over standard collaborative filtering or scaling-based approaches. Taken together, our findings suggest that preserving and exploiting the lineage graph is vital for guiding resource-efficient decisions in LLM development.

\section{Related Work}
\label{sec:related}

\paragraph{Performance Estimation via Scaling Laws.}
Early studies on large language models (LLMs) have shown that core metrics like cross-entropy loss exhibit approximate power-law relationships with key design factors such as model size, dataset scale, and overall computational budget~\citep{kaplan2020scalinglawsneurallanguage,hoffmann2022an,hernandez2022scalinglawsinterpretabilitylearning,gordon-etal-2021-data}. While these classical “scaling laws” provide valuable insights into how performance grows with parameters and tokens, their estimates typically assume a single model family or a narrowly defined training setup. Consequently, scaling relationships learned from one architecture (e.g., Transformer-based language models) may not directly generalize to another.

Recent work expands scaling laws beyond a single family. For instance, \citet{ye2023how} investigate whether past performance records from multiple LLM variants can predict new model settings; \citet{DBLP:journals/corr/abs-2401-04757} show that aggregated benchmark scores (e.g., BIG-Bench) follow smooth scaling trends across diverse model families, yet individual tasks are more difficult to predict. \citet{ruan2024observationalscalinglawspredictability} explore observational rather than purely experimental scaling, leveraging over a hundred pretrained LLMs to fit parametric curves without training each from scratch. These lines of research challenge the assumption that scaling laws transfer seamlessly across all model families. Indeed, \citet{choshen2025a} highlight large predictive errors when applying scaling trends derived from one family to a different, structurally distinct family. Overall, while scaling laws are fundamental to LLM research, their reliance on homogeneous or near-homogeneous settings can limit predictive accuracy, especially when addressing newly merged or fine-tuned models for which classical scaling metrics (e.g., total FLOPs or training tokens) are not strictly comparable.

\paragraph{Predicting Downstream Task Performance from Observational Data.}
A growing body of work aims to bypass some limitations of traditional scaling laws by using \emph{observational} or \emph{empirical} records from existing models. Rather than modeling only cross-entropy loss, these methods directly estimate how well a new or partially evaluated LLM will perform on downstream benchmarks. One common strategy is to reduce the high cost of comprehensive benchmarking by sampling only a small subset of instances. For example, \citet{polo2024tinybenchmarks}, \citet{kipnis2025metabench}, and \citet{pacchiardi2024instances} show that evaluating each new LLM on 100--600 carefully chosen samples can often predict that LLM’s accuracy on thousands of held-out benchmark items with minimal error. Such “tiny benchmarks” mitigate the expense of model-by-model, full-scale evaluations.

Beyond sample-efficient evaluation, another theme is to learn joint representations of both models and tasks. \citet{zhang-etal-2024-collaborative} combine historical LLM results with model and task descriptors (e.g., parameter size, instruction-tuning data) to improve prediction via matrix factorization or neural collaborative filtering. A similar philosophy underlies the works of \citet{ruan2024observationalscalinglawspredictability} and \citet{choshen2025a}, who incorporate both scaling metrics and observed performance to predict complex “emergent” or “agentic” behaviors of new models. Although these observational approaches can generalize better across different LLM architectures, they seldom exploit fine-grained \emph{lineage} information---such as which parent checkpoints were merged or how a model was fine-tuned---even though such lineage frequently constrains or correlates performance across tasks.

\paragraph{Routing, Ensemble Methods, and Embedding-Based Prediction.}
A complementary direction addresses how to \emph{select} or \emph{route} queries among multiple candidate LLMs that each has known strengths. In kNN-based routing~\citep{shnitzer2024large,hu2024routerbench}, for instance, the system uses prompt embeddings or input features to decide which model is most likely to excel on a particular query; \citet{lu-etal-2024-routing} similarly train routing functions based on reward signals. Such approaches implicitly rely on predicting individual models’ performance from prior records, so improved performance estimation methods can feed directly into these routing pipelines. Meanwhile, \citet{ong2025routellm} propose a factorization approach (combining matrix factorization with label-enhancement techniques) that learns a dynamic “router” capable of choosing between smaller vs.\ larger models at inference time to optimize cost-performance trade-offs. These routing-oriented works share a common challenge: how to reliably predict an LLM’s performance on a new input. While embedding-based or factor-based methods capture prompt similarity, they again treat each model as a separate entity without leveraging the \emph{relationship} between parent and child or among merged “soups.”

\paragraph{Motivation for Lineage-Aware Estimation.}
Merging or fine-tuning from existing checkpoints is pervasive in LLM development~\citep{pmlr-v162-wortsman22a-model-soup}. Two “child” models derived from the same “parent” frequently behave similarly across tasks, and naive averaging of parent checkpoints can sometimes preserve or even enhance performance in certain domains. Nevertheless, conventional scaling laws or standard matrix factorization rarely encode such genealogical constraints. As a result, performance estimates for newly merged or fine-tuned models often rely on expensive direct evaluations or broad assumptions. Our work addresses this gap by building on empirical performance-based frameworks and injecting explicit \emph{lineage} regularization, thereby offering a more reliable way to predict the behavior of newly derived models without exhaustive re-benchmarking.

\section{Method}


\textcolor{black}{
Our goal is to predict the performance of Large Language Models (LLMs) on specific tasks or benchmarks by leveraging partial observations and model relationships. In particular, we address the cold-start setting, where we estimate the performance of LLMs that have not yet been constructed or evaluated.
}

\subsection{Problem Statement}

Let \(\mathcal{M}\) be the set of LLMs and \(\mathcal{X}\) be the set of problems. Define \(z_{u, i} \in \{0,1\}\) as the evaluation score for problem \(i \in \mathcal{X}\) solved by LLM \(u \in \mathcal{M}\). Here, \(z_{u, i} = 1\) if the LLM correctly solves the problem, and 0 otherwise. For some LLM-problem pairs, their true scores are already known, and using these observed data, our goal is to accurately estimate the scores for unobserved pairs.

\subsection{Existing Methods for LLM Performance Prediction}

\paragraph{Historical-Performance-Based Methods.}
Performance-based prediction methods generally rely on historical performance data. These methods assume that if two models or problems exhibit similar observed performance, their unobserved performances will also be similar. Formally, they define a general function \(f(\cdot)\) that estimates performance using latent embeddings or features of models and instances:
\[
\hat{z}_{u,i} = f(m_u, x_i)
\]
where \(m_u \in \mathbb{R}^d\) and \(x_i \in \mathbb{R}^d\) are embeddings for the model \(u\) and instance \(i\), respectively. Specific examples include:

First, in the \textbf{Matrix Factorization (MF)} approach, the prediction is computed as
\[
\hat{z}_{u,i} = \sigma(m_u x_i^\top),
\]
where \( m_u \) and \( x_i \) are latent vectors representing the user and item, respectively, and \( \sigma \) is the sigmoid function.

Second, in the \textbf{Item Response Theory (IRT)} model, the prediction is given by
\[
\hat{z}_{u,i} = \sigma(a_i \theta_u - b_i),
\]
where \( \theta_u \) denotes the ability parameter of user \( u \), and \( a_i \) and \( b_i \) are the discrimination and difficulty parameters of item \( i \), respectively.

Finally, the \textbf{Neural Collaborative Filtering (NCF)} model estimates the interaction as
\[
\hat{z}_{u,i} = \sigma(\text{MLP}(m_u, x_i \mid \theta)),
\]
where \(\text{MLP}(\cdot)\) is a multi-layer perceptron parameterized by \(\theta\), applied to the concatenation of \( m_u \) and \( x_i \).

To learn these embeddings, methods typically minimize a binary cross-entropy (BCE) loss on observed performance scores, combined with an \(L_2\) regularization term:
\[
\mathcal{O}_{\text{BCE}} = -\frac{1}{|\Omega|} \sum_{(u,i) \in \Omega} \left[z_{u,i} \log(\hat{z}_{u,i}) + (1 - z_{u,i}) \log(1 - \hat{z}_{u,i})\right] + \lambda_{\text{L2}} (\|M\|_F^2 + \|X\|_F^2)
\]
where \(\Omega\) is the set of observed LLM-instance pairs, \(M\) and \(X\) are the matrices containing embeddings \(m_u\) and \(x_i\) as rows, respectively, and \(\|\cdot\|_F\) denotes the Frobenius norm.

The methods described above face significant limitations when dealing with new LLMs that have no observed performance data—the well-known cold-start problem in recommendation systems. This issue is particularly acute in LLM development, where developers must often make decisions about newly created models without having the resources to evaluate them extensively.

\paragraph{Factor-Enhanced Method.}
Further improvements can be achieved by explicitly incorporating model and task design factors, such as parameter size, dataset scale, and task characteristics, into the prediction~\citep{he2017neural,zhang-etal-2024-collaborative}. The enhanced NCF (\textbf{NCF with factors}) includes these additional factors as embedding vectors \(e_{v_i}, e_{v_j}\):
\[
\hat{s}_{ij} = \text{MLP}(p_i, q_j, e_{v_i}, e_{v_j}|\theta)
\]

Specifically, \citet{zhang-etal-2024-collaborative} considered model-related factors including Model Family (e.g., LLaMA 2, Pythia), Pretraining Dataset Size (tokens), Parameter Size, GPU Time, FLOPs, Context Window Size, Batch Size, Number of Layers, Number of Attention Heads, Key/Value Size, Bottleneck Activation Size, and Carbon Emission. They also included task-related factors such as Ability (e.g., reasoning), Task Family (e.g., ARC), Output Format (e.g., binary classification), and Few-Shot Setting.

Shapley value analysis indicated that among model factors, \emph{Pretraining Dataset Size}, \emph{Model Family}, and \emph{Batch Size} were most critical, while for task factors, \emph{Ability} was most influential. These results underscore that factors beyond traditional scaling laws significantly influence prediction accuracy.

Notably, in a homogeneous model set derived from the same model family, factors such as Parameter Size, Context Window Size, and Batch Size often remain identical or highly similar across models.

\subsection{Proposed: Lineage-Based Prediction}

We address the cold-start problem by leveraging an underutilized source of information: the \emph{lineage relationships} between LLMs. When a new model is derived through fine-tuning or merging existing models, its behavior is likely to be inherited from its "ancestors" or "parents." This intuition forms the foundation of our approach.
While factor-based methods (e.g. NCF with factors) attempt to transfer information indirectly by identifying models with similar characteristics (e.g., similar parameter sizes, training data, or architectural features), our lineage-based approach enables \emph{direct information transfer} along known derivation paths. This distinction is crucial: rather than inferring similarities from potentially noisy or incomplete metadata, we utilize the explicit knowledge of how models are derived from one another, capturing relationships that may not be apparent from factors alone.
To leverage lineage information effectively, we propose two distinct approaches. The first approach, Model Lineage Averaging, directly utilizes lineage relationships in their simplest form to make predictions. The second approach, Lineage-Regularized Matrix Factorization (LRMF), integrates lineage constraints into the collaborative filtering framework, allowing for more sophisticated modeling of performance transfer patterns between related models.

\paragraph{Proposed-1: Model Lineage Averaging}
Fine-tuned models and merged models inherently inherit the performance characteristics of their base models. We can estimate performance on an instance~$i$ for a new or sparsely evaluated LLM $u$ from the results of its \emph{lineage-related} (neighbor) LLMs:
\[
\hat{z}_{u,i} \;=\; \frac{1}{|\mathcal{N}_{\mathcal{M}}(u)|}
\sum_{u' \in \mathcal{N}_{\mathcal{M}}(u)} \;z_{u',i}.
\]
Here, the neighborhood $\mathcal{N}_{\mathcal{M}}(u)$ is defined based on lineage relationships. Specifically, if an LLM $u$ is derived from $v$ through a short lineage path (fine-tuning, merging, etc.), we treat $v$ as a neighbor of~$u$. This simple yet effective approach directly addresses the cold-start problem by transferring performance knowledge from established models to their derivatives.

\paragraph{Proposed-2: Lineage-Regularized Matrix Factorization (LRMF).}
While derived models fundamentally inherit capabilities from their base models, subtle differences emerge in practice during fine-tuning or merging processes. We account for these nuances by drawing inspiration from Collaborative Filtering approaches. Following the framework of Tag Informed Collaborative Filtering (TagiCoFi)~\citep{TagiCoFi}, which integrates additional information (e.g., user-generated tags) into the matrix factorization procedure to improve recommendation performance, we propose Lineage-Regularized Matrix Factorization (LRMF). LRMF extends traditional matrix factorization by explicitly encouraging similarity between embeddings of neighboring LLMs and neighboring tasks.

Specifically, let \(\mathbf{A}^{(\mathcal{M})}\) and \(\mathbf{A}^{(\mathcal{X})}\) be adjacency matrices indicating lineage connections among LLMs and similarity relationships among tasks, respectively. For models, \(\mathbf{A}^{(\mathcal{M})}\) directly encodes the lineage relationships: an edge exists between models if one is derived from the other through fine-tuning or merging. For example, if model $v$ is fine-tuned from model $u$, or if model $w$ is created by merging models $u$ and $v$, then \(\mathbf{A}^{(\mathcal{M})}_{u,v} = \mathbf{A}^{(\mathcal{M})}_{v,u} = 1\) and \(\mathbf{A}^{(\mathcal{M})}_{u,w} = \mathbf{A}^{(\mathcal{M})}_{v,w} = 1\).

We introduce regularization terms to ensure neighboring embeddings remain close. In particular, for models, the regularization term that enforces similarity between feature vectors of lineage-related models is given by:
\[
\mathcal{O}^{(\mathcal{M})}_2 = \frac{1}{2} \sum_{u,v} \mathbf{A}^{(\mathcal{M})}_{u,v} \|\mathbf{m}_u - \mathbf{m}_v\|^2.
\]

This term penalizes large differences between the embeddings of models that share a lineage connection. Expanding and simplifying this expression yields:
\[
\mathcal{O}^{(\mathcal{M})}_2 = \sum_{u} \mathbf{m}_u^\top \mathbf{m}_u\, \mathbf{D}^{(\mathcal{M})}_{u,u} - \sum_{u,v} \mathbf{m}_u^\top \mathbf{m}_v\, \mathbf{A}^{(\mathcal{M})}_{u,v} = \text{Tr}(\mathbf{M}^\top \mathbf{L}^{(\mathcal{M})}\mathbf{M}),
\]
where the diagonal degree matrix \(\mathbf{D}^{(\mathcal{M})}\) is defined by \(\mathbf{D}^{(\mathcal{M})}_{u,u} = \sum_v \mathbf{A}^{(\mathcal{M})}_{u,v}\) and the graph Laplacian is \(\mathbf{L}^{(\mathcal{M})} = \mathbf{D}^{(\mathcal{M})} - \mathbf{A}^{(\mathcal{M})}\).

Similarly, for tasks, we construct \(\mathbf{A}^{(\mathcal{X})}\) using cosine similarity between task embeddings, retaining the top \(k\) similar tasks, and define a corresponding regularization term \(\mathcal{O}^{(\mathcal{X})}_2 = \text{Tr}(\mathbf{X}^\top \mathbf{L}^{(\mathcal{X})}\mathbf{X})\).

The overall objective function is then formulated as:
\[
\mathcal{O} = \mathcal{O}_{\text{BCE}} + \lambda_{\mathcal{X}}\, \mathcal{O}^{(\mathcal{X})}_2 + \lambda_{\mathcal{M}}\, \mathcal{O}^{(\mathcal{M})}_2.
\]


Here, the regularization parameters $\lambda_{\mathcal{X}}$ and $\lambda_{\mathcal{M}}$ control the influence of instance similarity and model lineage constraints, respectively. A detailed analysis of these hyperparameters and their effects on model performance is provided in Appendix \ref{sec:appendix_hyperparam}. To optimize this objective function, we employ a stochastic gradient descent method with adaptive learning rates (Adam optimizer), which efficiently handles the non-convex nature of our formulation while providing fast convergence rates. This approach addresses data sparsity and significantly alleviates cold-start issues for newly derived models.

\section{Experiments}
\label{sec:experiments}
We evaluate the effectiveness of our proposed \textbf{Lineage-Based Prediction} methods through extensive empirical experiments using publicly available data from the Hugging Face Open LLM Leaderboard v2 \citep{open-llm-leaderboard-v2}. Our experiments focus on two critical scenarios encountered in real-world LLM development: (1) predicting performance among \textit{homogeneous} models (e.g., those derived from merging or fine-tuning within the same model family) and (2) predicting performance across more diverse, \textit{heterogeneous} models (e.g., different architectures or training methods).

\subsection{Experimental Setting}

We conducted large-scale experiments using evaluations from the Hugging Face Open LLM Leaderboard, focusing on 6 benchmark tasks (BBH \citep{suzgun-etal-2023-challenging}, GPQA \citep{rein2024gpqa}, IFEval \citep{zhou2023instructionfollowingevaluationlargelanguage}, MATH \citep{hendrycks2021measuring_MATH}, MMLU-Pro \citep{wang2024mmlupro}, MuSR \citep{sprague2024musr}) comprising 39 sub-benchmarks in total. Our dataset includes up to 2,934 publicly available models with partial or complete lineage information and 21,065 instances, though specific subsets were used depending on the experimental condition.

We compare three methods in our experiments:
\paragraph{Baseline: Neural Collaborative Filtering with Factors (NCF with factors)} 
Among our baselines, this is the only approach that estimates cold-start performance while explicitly leveraging historical results. We focus on cold-start methods that do not assume any target-model evaluations.\footnote{%
Methods such as \citet{polo2024tinybenchmarks} and \citet{kipnis2025metabench} assume the target model has been evaluated on at least a subset of instances, and are therefore inapplicable under our strict cold-start setting. Likewise, scaling-law approaches (e.g., \citet{ruan2024observationalscalinglawspredictability}) extrapolate from factors like parameter count or pretraining data volume—quantities that remain essentially unchanged under post-training or model merging—making them tend to be less informative for post-training or model merging for the present objective.}
We use the same network architecture as prior work \citep{zhang-etal-2024-collaborative}: a two-layer MLP with 128 hidden units per layer. Owing to practical constraints when collecting data from the leaderboard, we restrict model factors to three attributes: architecture type (50 categories; e.g., \texttt{qwen2forcasualllm}, \texttt{gemma2forcasuallm}), model type (three categories: fine-tuned, merged, and other—including base models), and parameter size. For task factors, we include benchmark identity (six majar benchmarks decomposed into 39 sub-benchmarks). All factors are embedded and concatenated with the model and instance embeddings before being fed to the MLP.

\paragraph{Proposed-1: Model-Lineage Averaging (MLA)}
We construct lineage relationships by extracting metadata from Hugging Face model cards. For fine-tuned models, we identify parent models via the \texttt{base model:finetune:} tag; for merged models, we use the \texttt{base model:merge:} tag and, when available, the \texttt{merge config.json} file. To ensure consistency across the corpus, we restrict model factors to those reliably present in the model-card JSON—``architecture type,'' ``model type,'' and ``parameter size.'' We omit factors such as dataset size or batch size because they are not reported consistently; in the homogeneous setting, such omitted factors are likely similar across models and thus have limited impact on our conclusions.


\paragraph{Proposed-2: Lineage-Regularized Matrix Factorization (LRMF)}
We construct the instance-similarity graph from semantic embeddings of the prompts. By default, we use \texttt{Snowflake/snowflake-arctic-embed-l-v2.0} to embed each prompt, compute cosine similarities, and connect each instance to its top-$k$ nearest neighbors ($k{=}20$ unless otherwise noted). The model-lineage graph reuses the lineage information described in the Model Lineage Averaging baseline, with an edge indicating a direct derivation between models. The model is implemented in PyTorch and optimized with Adam (learning rate $3\times 10^{-3}$). We train for at most 10{,}000 epochs with early stopping based on the validation set (patience of 100 epochs).

\paragraph{Hyperparameter Tuning}
Hyperparameters $\lambda_{L2}$, $\lambda_{\mathcal{M}}$, and $\lambda_{\mathcal{X}}$ were selected via grid search to maximize AUC-ROC on the development set, yielding the optimal values $\lambda_{L2}=10^{-5}$, $\lambda_{\mathcal{M}}=10^{-4}$, and $\lambda_{\mathcal{X}}=10^{-5}$ (see Appendix~\ref{sec:appendix_hyperparam} for details).

\paragraph{Evaluation Metric}
We evaluate each method using two complementary metrics that assess different aspects of prediction ability:
\begin{itemize}
\item \textbf{AUC-ROC (Individual Performance Estimation)}: Measures the ability to accurately estimate whether a single LLM's performance on a specific task exceeds a threshold, distinguishing correct (1) vs.\ incorrect (0) predictions at the instance level.
\item \textbf{Pearson Correlation (Relative Performance Ranking)}: Measures the ability to correctly preserve the relative performance hierarchy among multiple candidate LLMs on a given task, computed as the Pearson correlation coefficient between predicted and true average scores across benchmarks.
\end{itemize}
Pearson correlation is crucial for our goal of identifying the most promising model variants without exhaustive benchmarking, as it quantifies how well our method maintains the correct ranking of LLM performance. We report both metrics for each benchmark to provide a comprehensive assessment of individual score estimation accuracy and ranking fidelity.

\subsection{Experiment 1: Predicting Performance among Homogeneous Models} 

\paragraph{Scenario Setup} We first examine scenarios involving homogeneous models, i.e., those derived from a common base model through multiple fine-tunings or merges. This setting simulates practical use cases such as determining which model merges or fine-tunings are most promising.

We use 145 models derived from Qwen 2.5-7B, dividing them into 105 models for training, 20 for validation, and 18 for testing. (The detailed configuration of fine-tuned and merged models is listed in the Appendix.) 
We sample 1,000 instances from the leaderboard benchmarks and ensure that every model is both trained and evaluated on the same set of 1,000 instances.
This choice follows findings that performance on 100--600 samples reliably predicts overall benchmark performance \citep{polo2024tinybenchmarks,kipnis2025metabench}.
For this experiment, we repeat the sampling procedure five times with different random seeds and report the mean and standard deviation across these runs.


\vspace{-3mm}
\begin{figure*}[ht] 
\centering 
\begin{subfigure}[b]{0.49\textwidth} 
\centering 
\includegraphics[width=\textwidth]{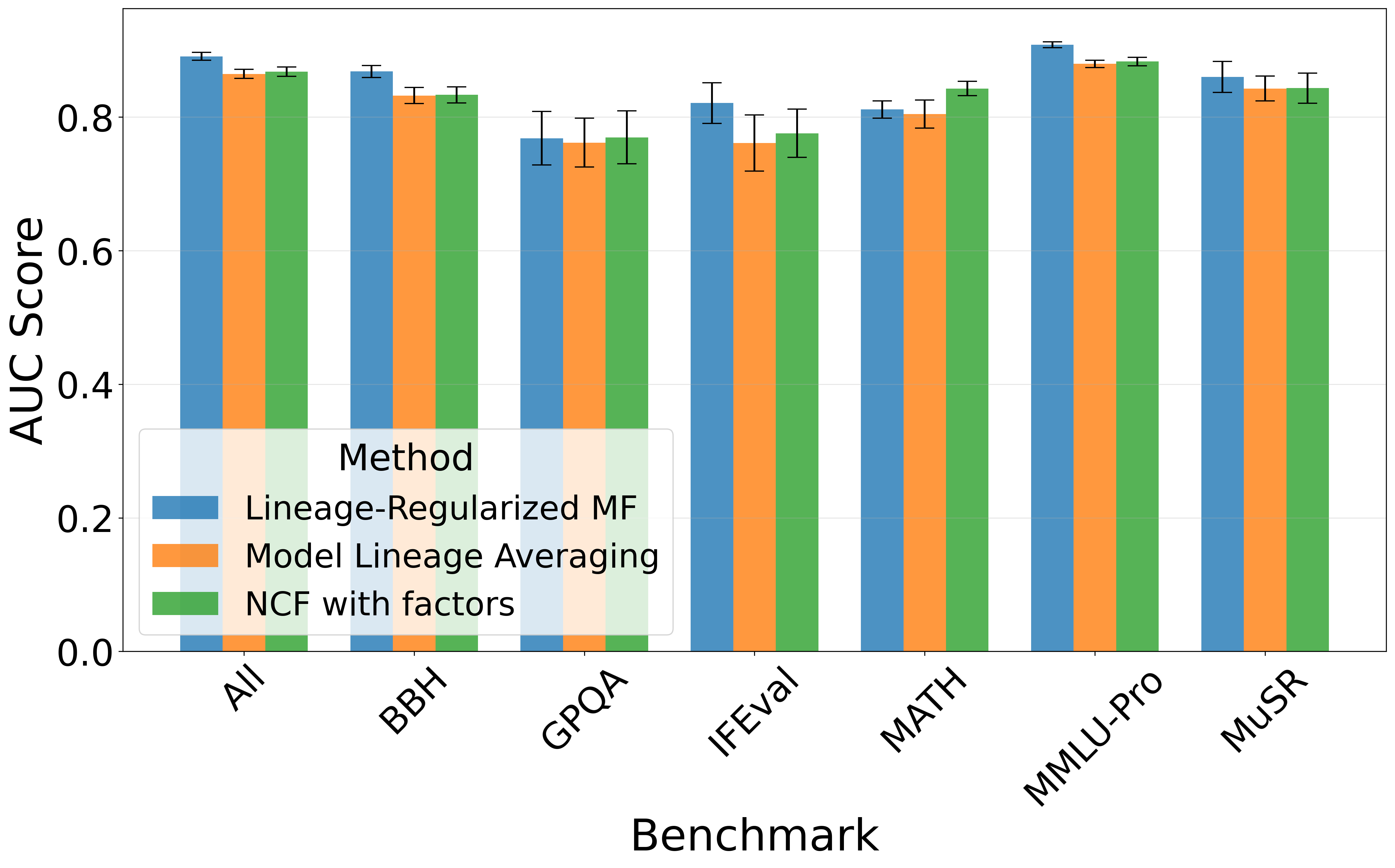} 
\caption{AUC-ROC} 
\label{fig:lambda_lineagermf_1_ex1} 
\end{subfigure} 
\hfill 
\begin{subfigure}[b]{0.49\textwidth} 
\centering 
\includegraphics[width=\textwidth]{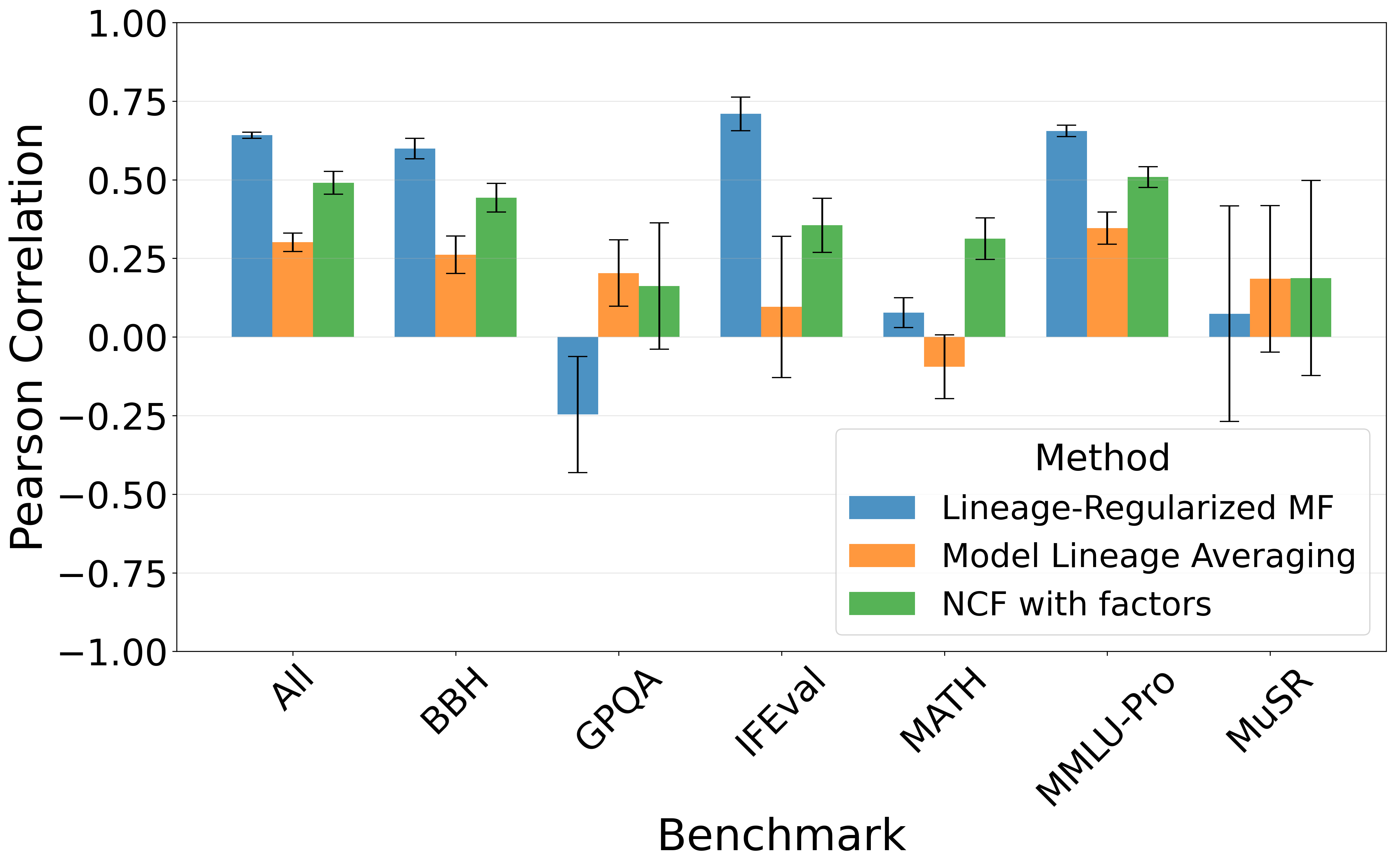} 
\caption{Pearson correlation} 
\label{fig:lambda_lineagermf_2_ex1} 
\end{subfigure} 
\caption{Performance evaluation results for a homogeneous set of models derived from Qwen 2.5-7B across various benchmarks. (a) AUC-ROC scores show that all methods achieve similar absolute performance estimation capability (0.83–0.86). (b) Pearson correlation reveals significant differences in relative performance ranking ability, with our proposed Lineage-Regularized MF achieving the highest overall correlation (0.642), substantially outperforming NCF with factors (0.490) and Model Lineage Averaging (0.301). This demonstrates the effectiveness of combining lineage information with collaborative filtering for preserving model performance hierarchies. Note the benchmark-specific patterns: LRMF excels on instruction-following (IFEval) and general reasoning tasks, while Model Lineage Averaging performs better on specialized knowledge (GPQA) where performance characteristics tend to be more directly inherited from parent models.} 
\label{fig:lambda_lineagermf_ex1} 
\end{figure*}

\vspace{-3mm}
\paragraph{Results}
Figure~\ref{fig:lambda_lineagermf_ex1} reports both AUC-ROC and Pearson correlation between the actual and predicted performance of each LLM across different benchmarks. All methods achieve high AUC-ROC scores (approximately 0.83–0.86), with our proposed Lineage-Regularized MF leading marginally. However, when evaluating relative performance via Pearson correlation, LRMF attains a substantially higher score (0.642) compared to NCF with factors (0.490) and Model Lineage Averaging (0.301). This discrepancy indicates that while existing methods can reasonably estimate absolute performance (as reflected by AUC-ROC), they struggle to distinguish the fine-grained performance ordering among multiple candidate models. In contrast, LRMF's incorporation of lineage information enables accurate prediction of model hierarchies beyond individual performance estimation.

Benchmark-specific results reveal meaningful contrasts in method effectiveness. On IFEval, which measures instruction-following capability, fine-tuning typically induces substantial behavioral changes. Consequently, methods relying primarily on ancestor scores—such as Model Lineage Averaging (MLA)—systematically underperform (correlation: 0.175), whereas LRMF, which exploits collaborative patterns beyond direct lineage, achieves substantially higher accuracy (correlation: 0.702). Conversely, on specialized knowledge benchmarks (GPQA) and complex reasoning tasks (MATH, MuSR), the Qwen-2.5-7B family demonstrates limited performance—most models solve very few problems—making relative ranking inherently challenging and resulting in lower correlations across all methods.

\subsection{Experiment 2: Predicting Performance among Heterogeneous Models}

\paragraph{Scenario Setup} To address realistic model merging scenarios, such as Frankenmerge \citep{pmlr-v162-wortsman22a-model-soup}, we examine performance prediction across diverse models with different architectures and parameter sizes. Specifically, we test whether our method remains effective compared to scaling-law-based approaches like NCF with factors, which typically benefit from heterogeneous training sets.

We train using 2,534 models, with 200 models each for validation and testing. As in Experiment 1, we sample 1,000 instances from the leaderboard benchmarks.

\begin{figure*}[ht] 
\centering 
\begin{subfigure}[b]{0.49\textwidth} 
\centering 
\includegraphics[width=\textwidth]{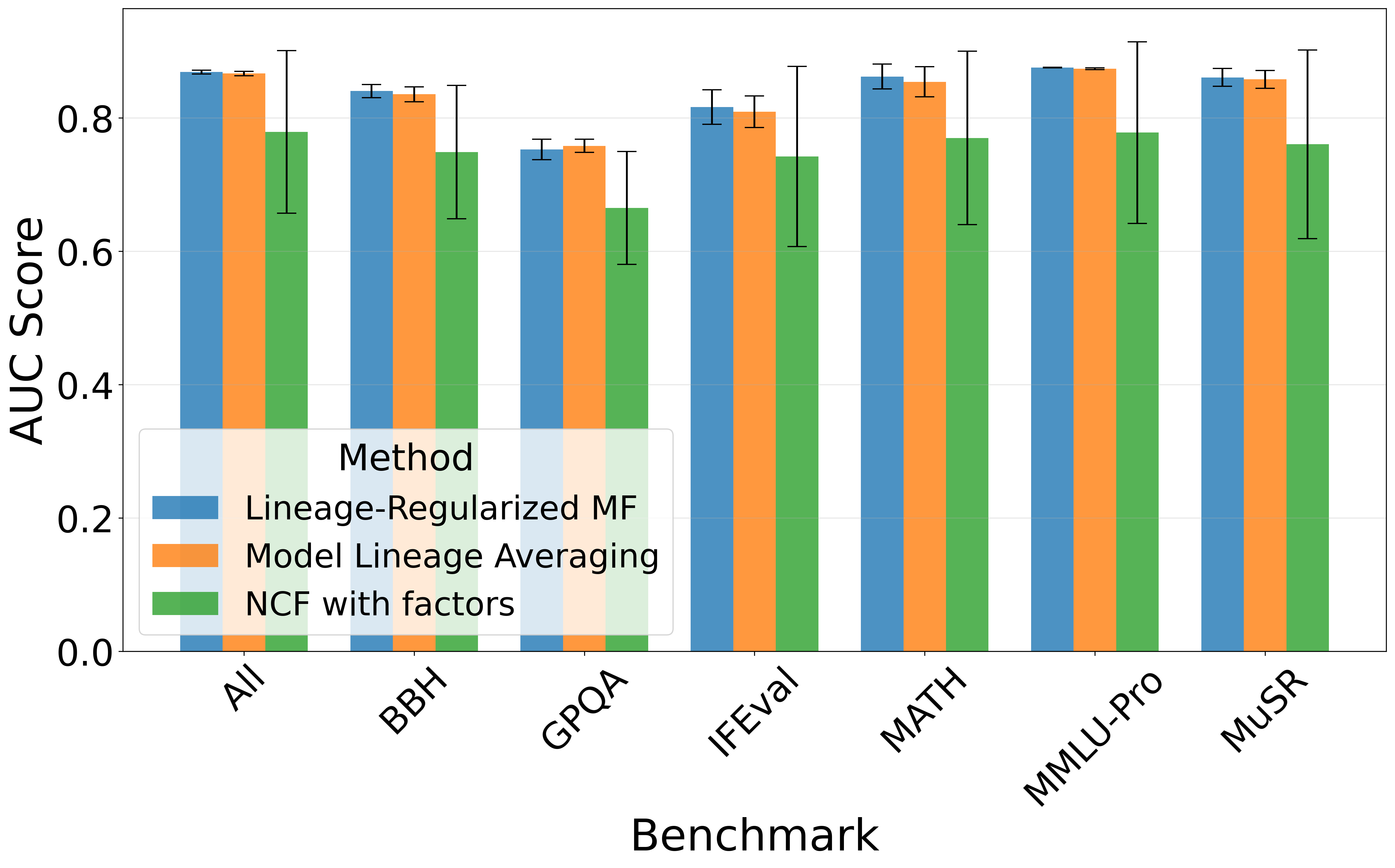} 
\caption{AUC-ROC (absolute evaluation)} 
\label{fig:lambda_lineagermf_1_ex2} 
\end{subfigure} 
\hfill 
\begin{subfigure}[b]{0.49\textwidth} 
\centering 
\includegraphics[width=\textwidth]{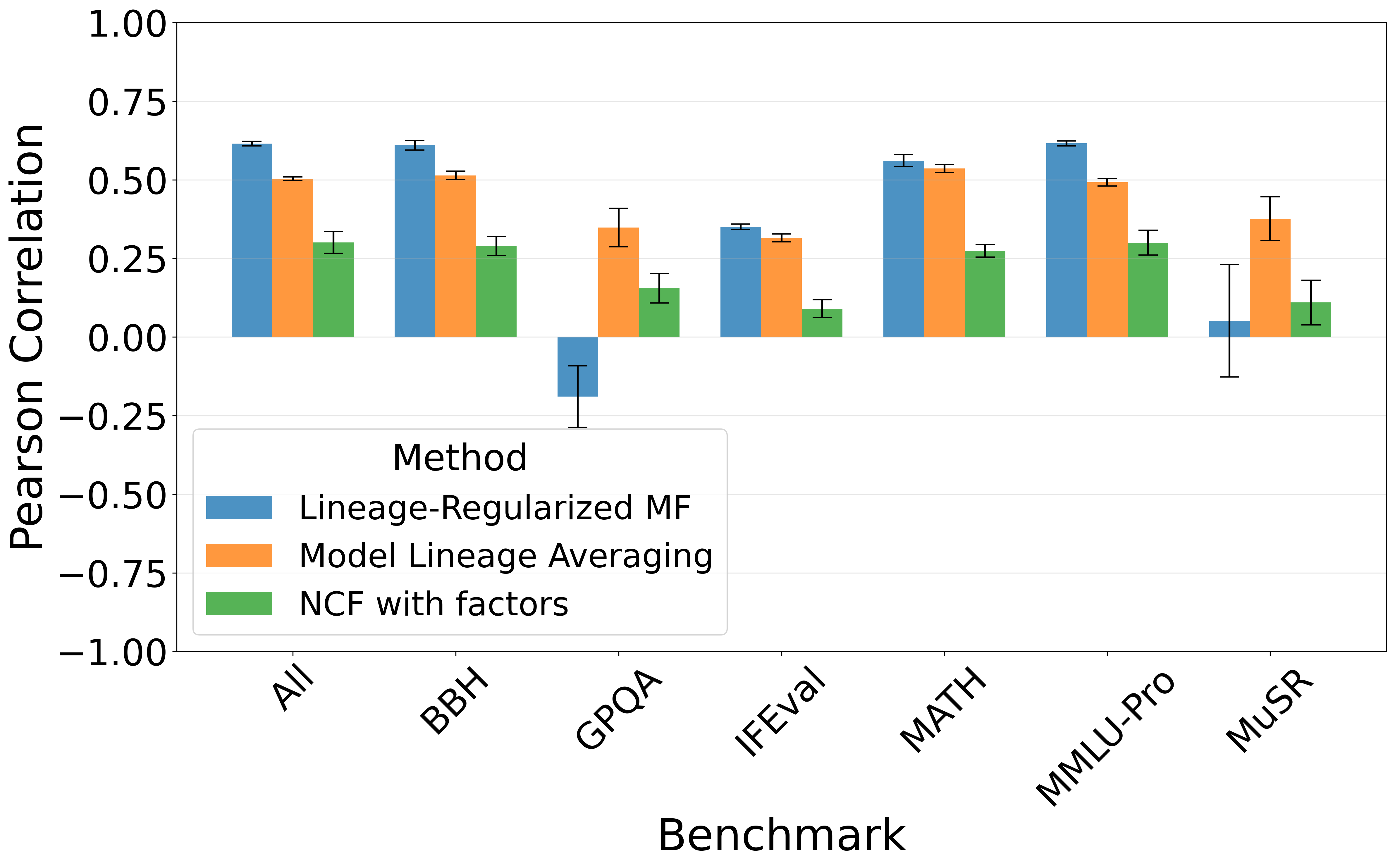} 
\caption{Pearson correlation (relative evaluation)} 
\label{fig:lambda_lineagermf_2_ex2} 
\end{subfigure} 
\caption{Performance evaluation results for heterogeneous models from the Hugging Face Open LLM Leaderboard. (a) AUC-ROC scores remain similar across methods. (b) Pearson correlation shows LRMF achieving the highest overall correlation (0.615), demonstrating effectiveness across diverse architectures and training strategies.} 
\label{fig:lambda_lineagermf_ex2} 
\end{figure*}

\paragraph{Results}
Figure~\ref{fig:lambda_lineagermf_ex2} reports both AUC-ROC and Pearson correlation in the heterogeneous setting. All methods again exhibit similarly high AUC-ROC ($\approx$ 0.84–0.88), while Pearson correlation shows clear separation: LRMF reaches 0.615, Model Lineage Averaging 0.504, and NCF with factors 0.301. This mirrors the homogeneous results—while traditional approaches can estimate absolute accuracy, lineage-aware modeling in LRMF most reliably captures the relative ordering of diverse models.

Compared with Experiment~1, NCF with factors shows a relative decline. This underperformance likely reflects the factor-importance bias observed in prior work~\citep{zhang-etal-2024-collaborative}, where SHAP analyses indicated that coarse factors (e.g., model family) dominate predictions. When such strong explanatory factors exist, the model tends to collapse predictions within each family and fails to resolve within-family distinctions—behavior that may inflate correlation in homogeneous settings but does not transfer to heterogeneous ecosystems.

Benchmark-specific trends are largely consistent with Experiment~1. Lineage-Regularized Matrix Factorization continues to excel on BBH, IFEval, MATH, and MMLU-Pro, whereas Model Lineage Averaging remains stronger on GPQA and MuSR. This stable pattern across settings reinforces our view of how distinct capabilities propagate along model lineage. Notably, MLA becomes relatively stronger in this heterogeneous scenario, and the gap between LRMF and MLA narrows. This suggests that greater model diversity requires the latent space to encode more complex relationships; in this regime, direct lineage information serves as a comparatively stable signal, whereas embedding-based methods can struggle to represent nuanced differences across architectures and training recipes within a unified latent space.



\subsection{Experiment 3: Routing}

To evaluate the practical applicability of our performance predictions, we consider an instance-level routing scenario where each input must be dynamically assigned to the LLM expected to perform best. We compare five routing strategies: (i) LRMF-based Routing (ours), (ii) Model Lineage Averaging (MLA)-based Routing, (iii) Neural Collaborative Filtering (NCF) with factors-based Routing, (iv) Random Routing, and (v) Best Model, which always assigns all instances to the single model with the highest average true performance.

\begin{figure*}[ht] 
\centering 
\begin{subfigure}[b]{0.49\textwidth} 
\centering 
\includegraphics[width=\textwidth]{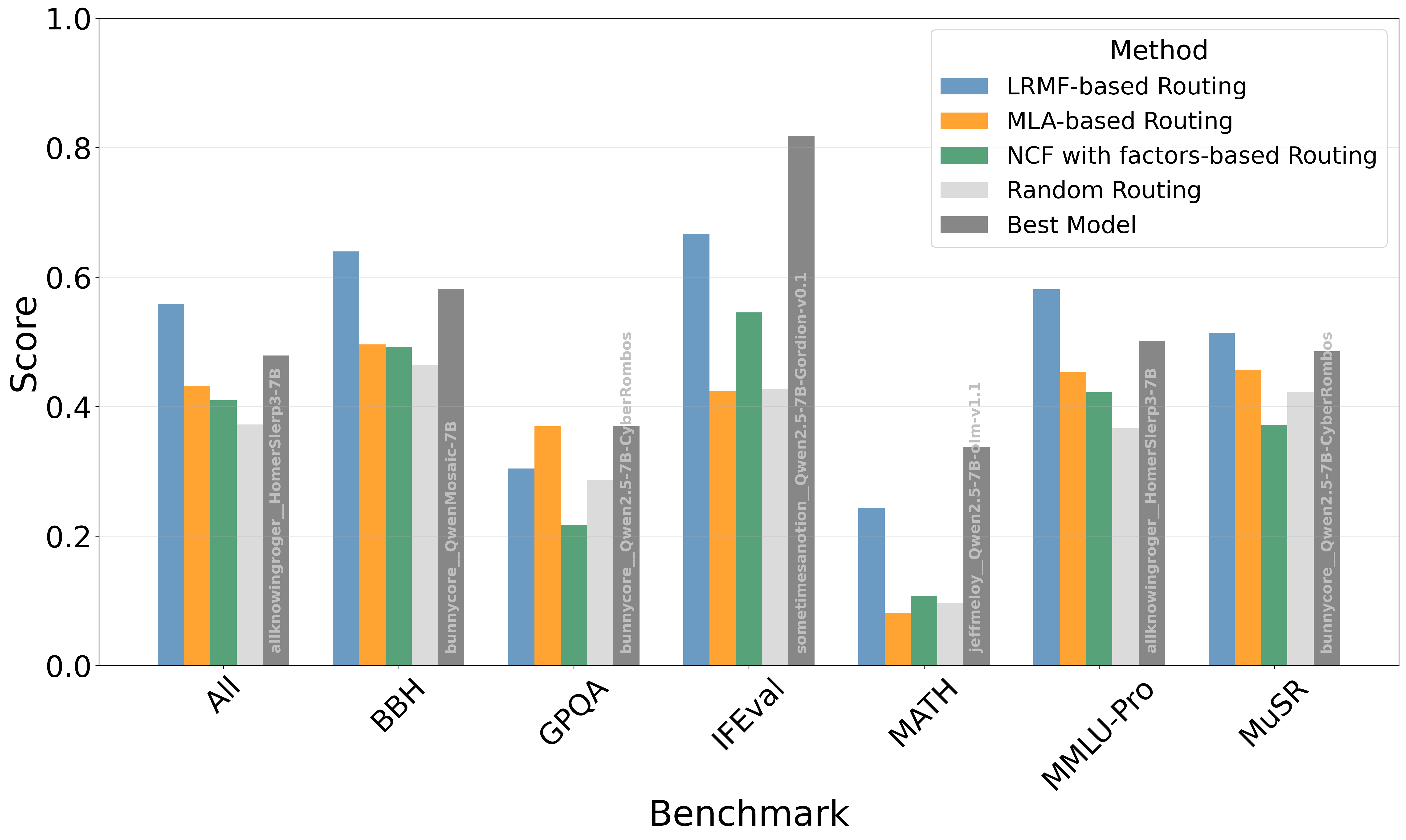} 
\caption{Routing performance in \emph{homogeneous} setting} 
\label{fig:qwen_routing_scores_comparison_all_methods} 
\end{subfigure} 
\hfill 
\begin{subfigure}[b]{0.49\textwidth} 
\centering 
\includegraphics[width=\textwidth]{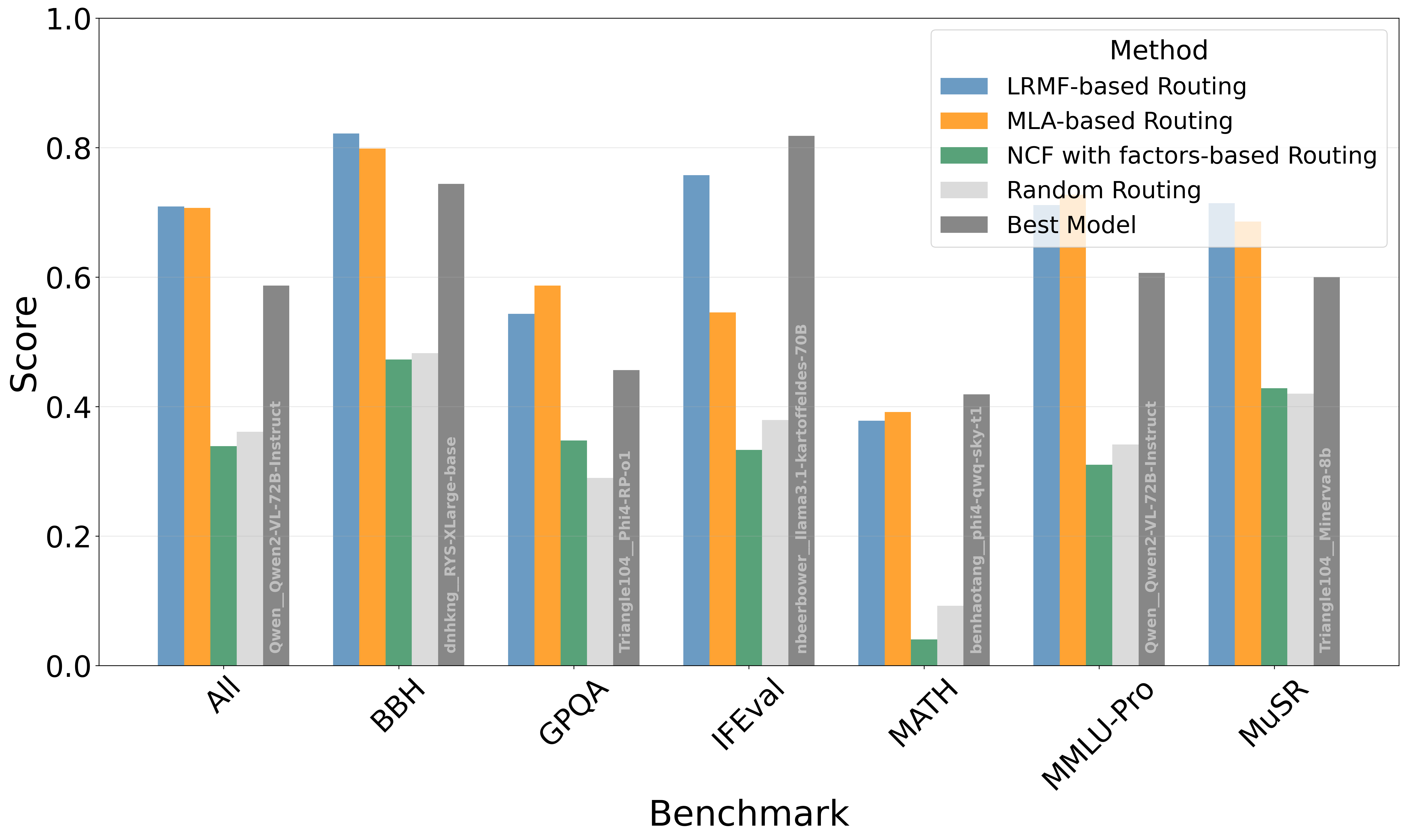} 
\caption{Routing performance in \emph{heterogeneous} setting} 
\label{fig:all_routing_scores_comparison_all_methods} 
\end{subfigure}  
\caption{Instance-level routing performance across benchmarks in (a) homogeneous and (b) heterogeneous settings. Each test instance is dynamically assigned to the model with the highest predicted score according to different methods. We compare LRMF-based routing against MLA, NCF with factors, Random assignment, and Best Model (always using the single top-performing model per benchmark). LRMF routing consistently outperforms all baselines in both settings, demonstrating that incorporating lineage information enables more effective dynamic model selection that exceeds the performance of using a single optimal model for all instances.}
\label{fig:routing} 
\end{figure*}

As shown in Figure~\ref{fig:routing}, LRMF-based routing consistently outperforms all baselines in both homogeneous and heterogeneous settings. Crucially, our method exceeds the Best Model baseline—the minimum threshold for effective instance-level routing—demonstrating that dynamic model assignment based on predicted performance can improve overall benchmark results beyond using a single optimal model.

These findings indicate that explicit modeling of lineage relationships substantially improves model selection in practical routing scenarios, enabling more effective dynamic assignment of inputs to maximize overall performance.

\section{Conclusion}
\label{sec:conclusion}
Our work demonstrates that lineage relationships---the explicit parent-child connections between derived LLMs---significantly improve performance prediction accuracy. Across experiments with 2,934 models and six major benchmarks, our \textbf{Lineage-Regularized Matrix Factorization (LRMF)} framework achieved up to \textbf{0.15--0.30} increases in correlation with actual performance by effectively modeling ancestral connections between models.

Our analysis revealed that while instruction-following abilities undergo substantial changes during fine-tuning or merging, specialized knowledge and complex reasoning capabilities remain more closely tied to parent models. These insights have practical implications for LLM development, allowing researchers to efficiently explore model variants without exhaustive re-benchmarking, thus reducing computational costs and accelerating development cycles.

Limitations include reduced effectiveness for novel architectures without clear lineage and uniform treatment of all lineage connections regardless of modification type. Future work could explore weighted connections and deeper analysis of how specific fine-tuning techniques affect capability inheritance.

As the LLM ecosystem continues expanding through derivatives rather than training from scratch, tracking lineage information becomes increasingly valuable. Our work shows that understanding ``where a model comes from'' is just as important as knowing ``what it is'' when predicting LLM performance.

\bibliography{colm2025_conference}
\bibliographystyle{colm2025_conference}

\appendix




\section{Detailed Model Statistics} 
\label{sec:appendix_model_stats}

We present more extensive information about the models used in each experiment. This includes both the partition splits (train / dev / test) and a breakdown by model type: whether a given model checkpoint is (1) a \emph{Fine-Tuned (FT)} model, derived from a single parent model via continued training on additional or specialized datasets; (2) a \emph{Merged} model, created through parameter-level averaging or “souping” of multiple parent models; or (3) an \emph{Other} model, which encompasses base models released without any further fine-tuning or merging, as well as models for which lineage information is incomplete or unavailable.

\subsection{Experiment 1: Homogeneous Models (Qwen-2.5-7B Family)} Table~\ref{tab:homo_stats} lists all 143 models in the Qwen 2.5-7B family used in the homogeneous setting, along with their assigned partition (105 for training, 20 for development, 18 for testing) and their lineage type (Merged / FT / Other). This family traces back to the publicly available Qwen 2.5-7B checkpoint, with each descendant model created via single or multiple merges and/or domain-specific fine-tuning steps.

\begin{figure}[ht]
    \centering
    \includegraphics[width=0.95\textwidth]{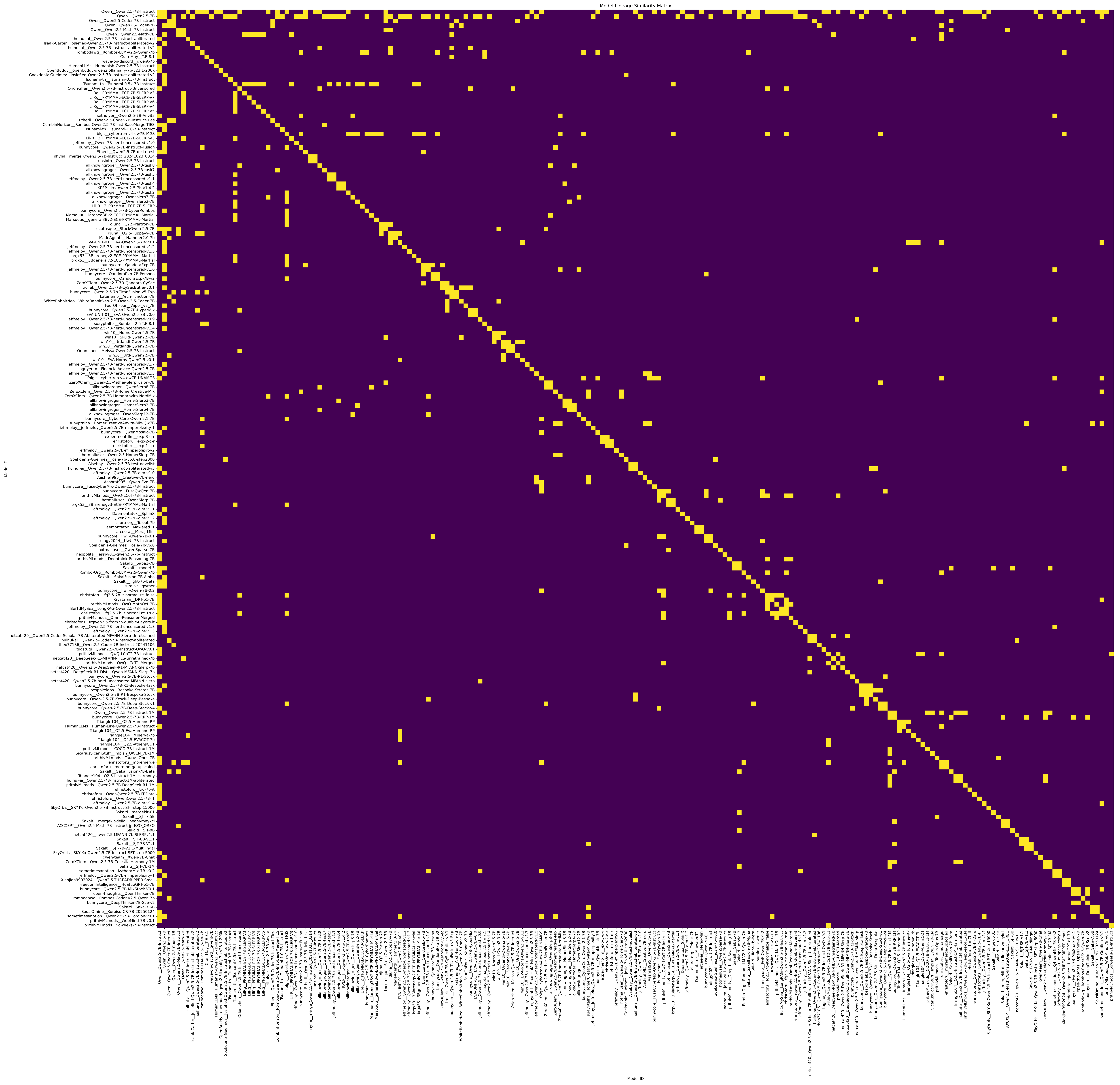}
    \caption{Lineage network visualization among the Qwen 2.5-7B models. The most densely connected model is \texttt{Qwen-2.5-7B-Instruct}.}
    \label{fig:lineage_heatmap_qwen}
\end{figure}

\paragraph{Lineage Analysis of Qwen 2.5-7B Family} We investigated the lineage distribution within this homogeneous group, discovering that the number of lineage connections roughly follows a logarithmic distribution (Figure~\ref{fig:num_lineage_distribution_qwen}). Furthermore, we visualized the complete lineage network among the Qwen models (Figure~\ref{fig:lineage_heatmap_qwen}), revealing that the model with the most lineage connections is \texttt{Qwen-2.5-7B-Instruct}.

\subsection{Experiment 2: Heterogeneous Models (All Models)} In the heterogeneous scenario (Experiment 2), we used a total of 2,934 models from the Hugging Face Open LLM Leaderboard that included partial or complete lineage metadata. These were split into 2,534 for training, 200 for development, and 200 for testing. Due to the diversity in architectures, sizes, merge strategies, and fine-tuning procedures, we do not list each model name. Instead, Table\ref{tab:hetero_stats} summarizes the approximate proportions of base, merged, and fine-tuned models.

\begin{figure*}[ht]
\centering
\begin{subfigure}[b]{0.49\textwidth}
    \centering
    \includegraphics[width=\textwidth]{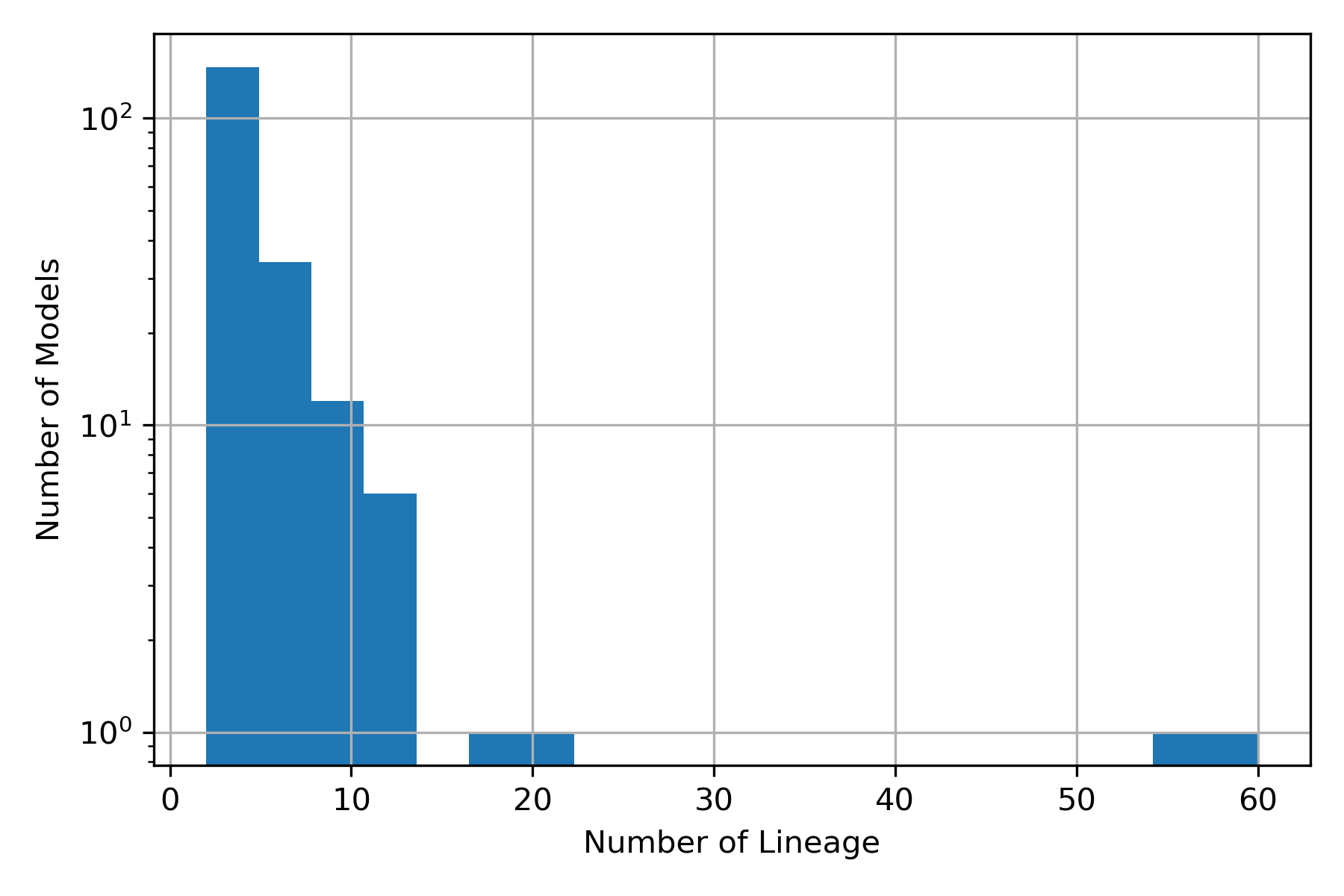}
    \caption{Qwen 2.5-7B family models}
    \label{fig:num_lineage_distribution_qwen}
\end{subfigure}
\hfill
\begin{subfigure}[b]{0.49\textwidth}
    \centering
    \includegraphics[width=\textwidth]{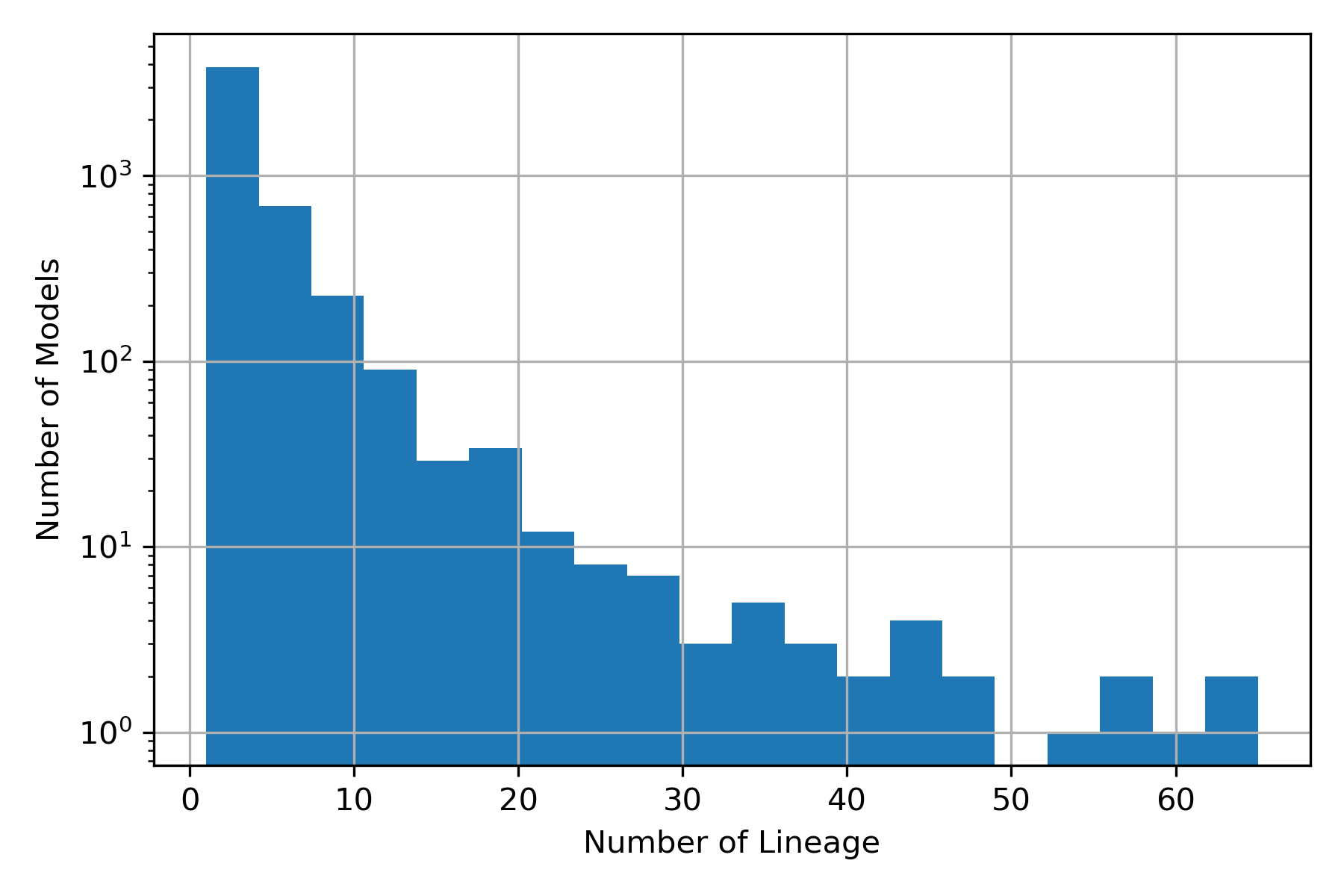}
    \caption{All models}
    \label{fig:num_lineage_distribution_all}
\end{subfigure}
\caption{Distribution of lineage connections among models. Both distributions approximate a logarithmic trend, indicating many models have relatively few lineage connections, with only a minority having extensive lineages. (a) Homogeneous Qwen 2.5-7B family. (b) All models from the Hugging Face Open LLM Leaderboard.}
\label{fig:lineage_distribution}
\end{figure*}

\paragraph{Lineage Analysis of Heterogeneous Models} We also conducted a lineage analysis across the full heterogeneous set, uncovering again a roughly logarithmic distribution in the number of lineage connections (Figure~\ref{fig:num_lineage_distribution_all}). Approximately 60\% of these models are either fine-tuned or merged variants, suggesting lineage information is widely applicable for modeling relationships and improving predictions across a broad spectrum of publicly available models.

\begin{table}[ht]
\centering
\caption{Composition of the \emph{homogeneous} model set (Qwen 2.5-7B family) used in Experiment~1.}
\label{tab:homo_stats}
\begin{tabular}{lcccc}
\toprule
\textbf{Partition} & \textbf{Total} & \textbf{FT Models} & \textbf{Merged Models} & \textbf{Other (including Base)} \\
\midrule
Train & 105 & 45 (42.9\%) & 60 (57.1\%) & 0 (0.0\%) \\
Dev   & 20  & 8 (40.0\%)  & 12 (60.0\%) & 0 (0.0\%) \\
Test  & 18  & 9 (50.0\%)  & 9 (50.0\%)  & 0 (0.0\%) \\
\bottomrule
\end{tabular}
\end{table}

\begin{table}[h]
\centering
\caption{Composition of the \emph{heterogeneous} model set (All models from the Hugging Face Open LLM Leaderboard) used in Experiment~2.}
\label{tab:hetero_stats}
\begin{tabular}{lcccc}
\toprule
\textbf{Partition} & \textbf{Total} & \textbf{FT Models} & \textbf{Merged Models} & \textbf{Other (including Base)} \\
\midrule
Train & 2,534 & 856 (33.8\%) & 707 (27.9\%) & 971 (38.3\%) \\
Dev   & 200   & 56 (28.0\%)  & 54 (27.0\%)  & 90 (45.0\%) \\
Test  & 200   & 54 (27.0\%)  & 66 (33.0\%)  & 80 (40.0\%) \\
\bottomrule
\end{tabular}
\end{table}

\clearpage

\section{Hyperparameter Tuning and Ablation Study}
\label{sec:appendix_hyperparam}

Our experiments involve performance data collected from six benchmark tasks provided by the Hugging Face Open LLM Leaderboard, totaling approximately 21,000+ instances. For each instance, we have binary outcomes indicating whether a given LLM correctly solved the instance or not. The task is thus formulated as predicting these binary instance-level outcomes for model-instance pairs for which no evaluation is available.

\subsection{Experimental Protocol}

All hyperparameter optimization was performed on the validation set using two complementary metrics:
\begin{itemize}
    \item \textbf{Primary metric}: Instance-level AUC-ROC, which captures the model's ability to distinguish between correct and incorrect predictions
    \item \textbf{Secondary metric}: Pearson correlation coefficient, which measures how well the method preserves relative performance superiority relationship among models
\end{itemize}

For all methods, we performed grid searches over relevant regularization parameters and selected the configuration that maximized AUC-ROC on the development set. Additionally, we conducted comprehensive ablation studies using Pearson correlation to better understand the contribution of each component.

\subsection{Lineage-Regularized MF}

Lineage-Regularized MF extends the conventional matrix factorization by including additional Laplacian regularization terms:
\[
\lambda_{\mathcal{M}}\,\mathrm{Tr}\!\left(M^\top L^{(\mathcal{M})} M\right) + \lambda_{\mathcal{X}}\,\mathrm{Tr}\!\left(X^\top L^{(\mathcal{X})} X\right),
\]
where \(L^{(\mathcal{M})}\) and \(L^{(\mathcal{X})}\) are the graph Laplacians for models and instances, respectively, and an \(L_2\) penalty with coefficient \(\lambda_{\mathrm{L2}}\) is also applied. The hyperparameters \(\lambda_{\mathrm{L2}}, \lambda_{\mathcal{M}}, \lambda_{\mathcal{X}}\) were tuned over the range \(\{10^{-9}, 10^{-8}, \ldots, 10^{0}\}\) using the development set.

\subsubsection{AUC-ROC Analysis}
\begin{figure*}[ht] 
\centering 
\begin{subfigure}[b]{0.49\textwidth} 
\centering 
\includegraphics[width=\textwidth]{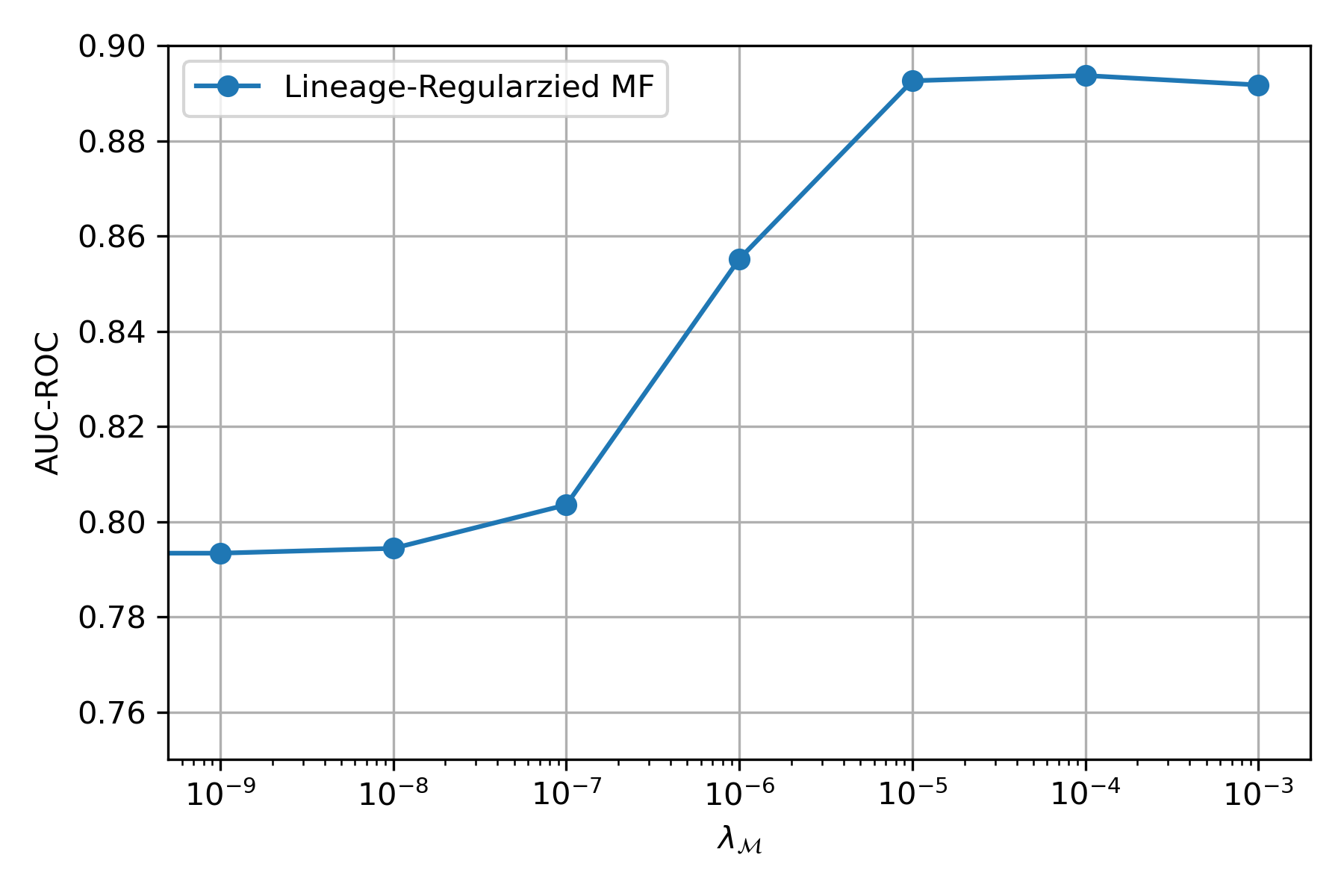} 
\caption{Varying $\lambda_{\mathcal{M}}$ (model lineage) with fixed $\lambda_{\mathcal{X}}=10^{-5}$} 
\label{fig:lambda_lineagermf_1} 
\end{subfigure} 
\hfill 
\begin{subfigure}[b]{0.49\textwidth} 
\centering 
\includegraphics[width=\textwidth]{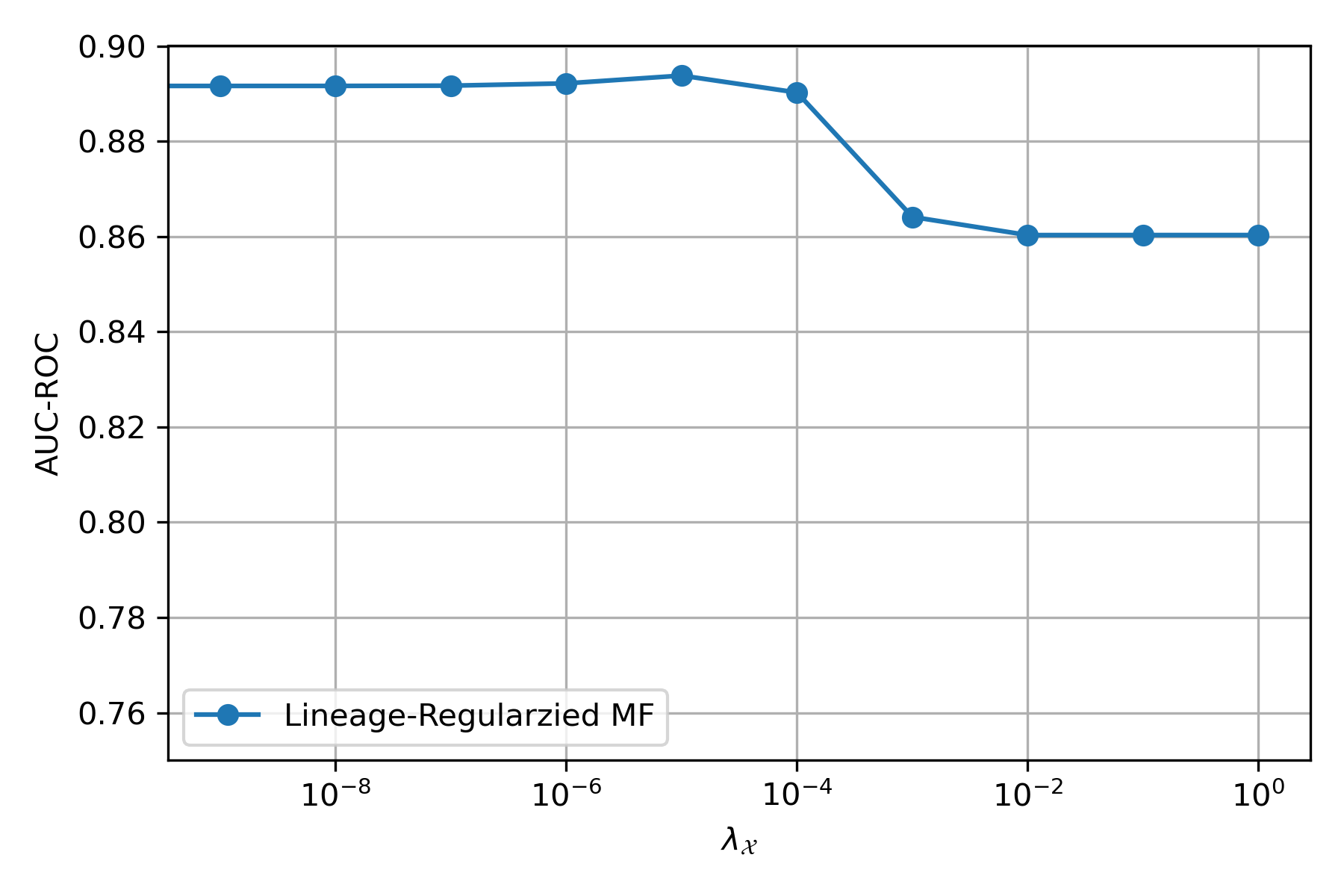} 
\caption{Varying $\lambda_{\mathcal{X}}$ (instance similarity) with fixed $\lambda_{\mathcal{M}}=10^{-4}$} 
\label{fig:lambda_lineagermf_2} 
\end{subfigure} 
\caption{Development-set AUC-ROC of Lineage-Regularized MF as a function of the lineage regularization parameter \(\lambda_{\mathcal{M}}\) and the instance regularization parameter \(\lambda_{\mathcal{X}}\). Each plot fixes one parameter while sweeping the other. Optimal performance is achieved with moderate regularization (\(\lambda_{\mathcal{M}} \approx 10^{-4}\), \(\lambda_{\mathcal{X}} \approx 10^{-5}\), ).} 
\label{fig:lambda_lineagermf} 
\end{figure*}

Figure~\ref{fig:lambda_lineagermf} illustrates the effect on AUC-ROC when varying \(\lambda_{\mathcal{M}}\) (lineage regularization) and \(\lambda_{\mathcal{X}}\) (instance similarity regularization), while fixing \(\lambda_{\mathrm{L2}}=10^{-5}\). 

For the model lineage regularization parameter \(\lambda_{\mathcal{M}}\), we observe a clear improvement as the regularization strength increases, with AUC-ROC rising from approximately 0.795 at \(\lambda_{\mathcal{M}} = 10^{-9}\) to a peak of approximately 0.895 at \(\lambda_{\mathcal{M}} = 10^{-4}\), demonstrating the critical importance of lineage information for performance prediction.

For the instance similarity regularization parameter \(\lambda_{\mathcal{X}}\), we observe that performance remains highly stable (AUC-ROC $\approx$ 0.895) for values at \(10^{-5}\), and then decreases to approximately 0.86 when \(\lambda_{\mathcal{X}} \geq 10^{-3}\), indicating over-regularization.

These results indicate that model lineage regularization (\(\lambda_{\mathcal{M}}\)) has a more pronounced impact on performance, with optimal values around \(10^{-4}\), while instance similarity regularization (\(\lambda_{\mathcal{X}}\)) shows broader stability across multiple orders of magnitude.

\subsubsection{Comprehensive Ablation Study}

\begin{figure}[ht]
    \centering
    \includegraphics[width=0.80\textwidth]{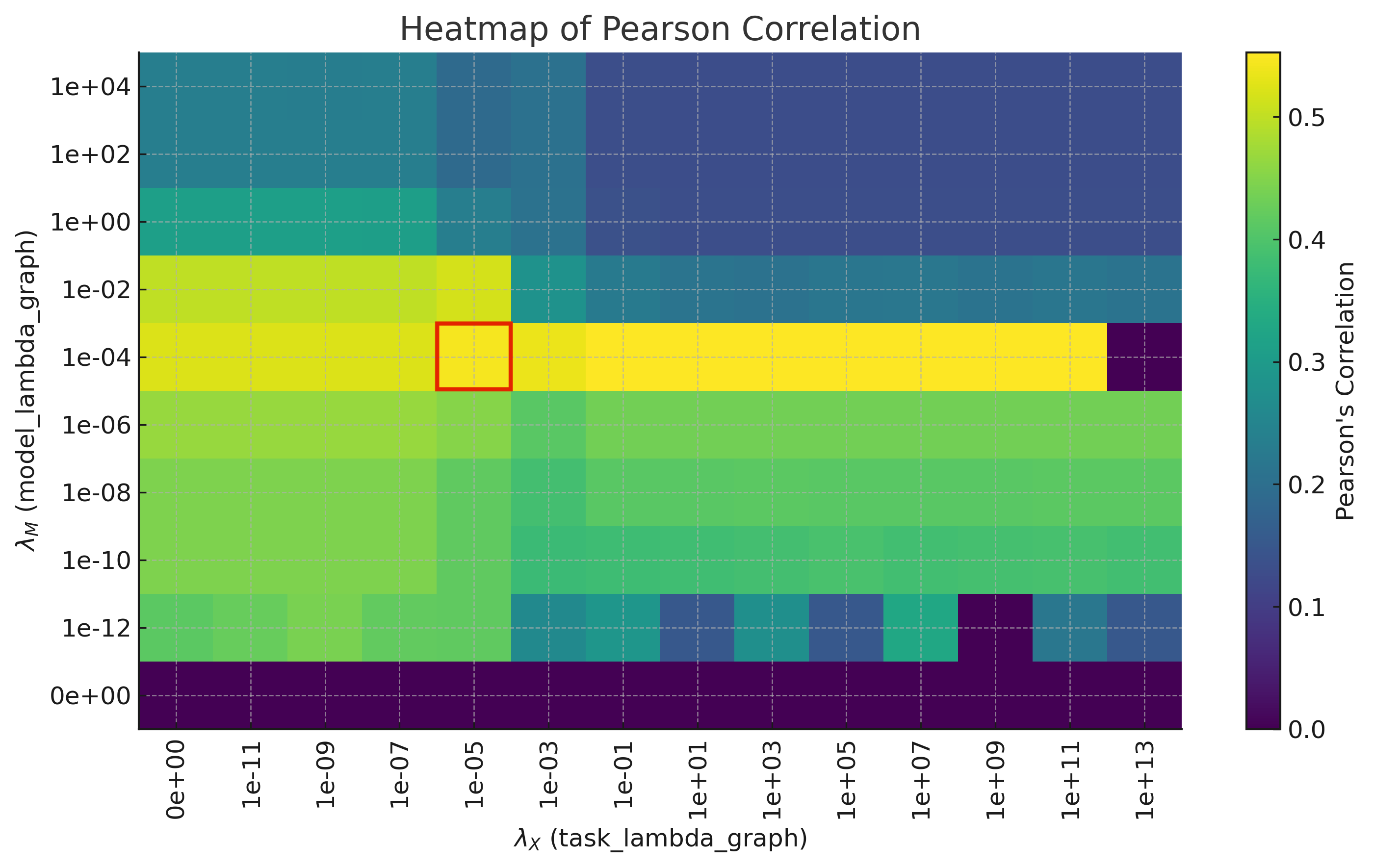}
    \caption{Heatmap of Pearson correlation coefficients for all combinations of $\lambda_{\mathcal{M}}$ (model lineage) and $\lambda_{\mathcal{X}}$ (task/instance similarity) regularization parameters. The red box highlights the optimal region. Note that $\lambda_{\mathcal{M}} = 0, \lambda_{\mathcal{X}} = 0$ (bottom-left) corresponds to standard MF and fails completely in cold-start prediction (correlation $\approx 0$). The visualization clearly shows that model lineage regularization ($\lambda_{\mathcal{M}} > 0$) is the primary driver of performance improvement.}
    \label{fig:ablation}
\end{figure}

To more thoroughly understand the interaction between regularization components and their individual contributions, we conducted an extensive ablation study measuring Pearson correlation coefficients across the full hyperparameter space (Figure~\ref{fig:ablation}).

\textbf{Critical observations from the ablation study:}
\begin{enumerate}
    \item \textbf{Baseline failure in cold-start}: When both $\lambda_{\mathcal{M}} = 0$ and $\lambda_{\mathcal{X}} = 0$ (equivalent to standard Matrix Factorization/NCF), the method completely fails in cold-start scenarios with correlation $\approx 0$, confirming that vanilla collaborative filtering cannot handle unseen models.
    
    \item \textbf{Model lineage as the dominant factor}: Setting $\lambda_{\mathcal{M}} > 0$ while keeping $\lambda_{\mathcal{X}} = 0$ achieves Pearson correlation coefficients of $0.2$--$0.5$, demonstrating that model lineage alone provides substantial predictive power even without instance similarity information. This shows that most of the correlation improvement stems from the model lineage regularization term.
    
    \item \textbf{Synergistic effects}: The optimal configuration ($\lambda_{\mathcal{M}} \approx 10^{-4}$, $\lambda_{\mathcal{X}} \approx 10^{-5}$) achieves correlation $> 0.5$, with the heatmap revealing a clear ``sweet spot'' (highlighted by the red box) where both regularization terms work synergistically.
    
    \item \textbf{Robustness of model lineage}: The correlation remains stable across a wide range of $\lambda_{\mathcal{M}}$ values ($10^{-6}$ to $10^{11}$), indicating that lineage information is inherently robust and does not require precise tuning.
\end{enumerate}

These ablation results validate our core hypothesis that explicit lineage relationships provide crucial signals for performance prediction, with the model lineage regularization term contributing the majority of the improvement. Once the best hyperparameter values are identified on the development set, we retrain Lineage-Regularized MF on the combined training and development sets and report the final performance on the test set.

\subsection{Ablation: Instance-graph construction}\label{app:ablation_instance_graph}
\paragraph{Embedding backbones.}
We evaluated several widely used embedding models for constructing the instance-similarity graph—\texttt{Snowflake/snowflake-arctic-embed-l-v2.0}, \texttt{infloat/e5-mistral-7b-instruct}, and \texttt{all-mpnet-base-v2}. For each model, we embedded the prompt text, computed cosine similarities, and formed a $k$-NN graph. Across benchmarks, we observed no material differences in downstream metrics (AUC-ROC and Pearson correlation), and the relative ordering of methods remained stable. Given this robustness and for consistency, we use \texttt{Snowflake/snowflake-arctic-embed-l-v2.0} in the main results.

\paragraph{Neighborhood size $k$.}
We swept $k \in \{2,5,10,20,50,100\}$ on the development set, selecting $k$ by AUC-ROC. Performance was largely insensitive to $k$ across tasks; we therefore fix $k{=}20$ in all reported experiments as a stable sparsity–connectivity trade-off.

\section{Robustness to Lineage Noise}
To quantitatively assess the impact of noise/incompleteness in lineage data, we conducted additional experiments by randomly modifying the lineage information (either adding or removing lineage links) in the \emph{heterogeneous} setting. Notably, we observed distinct responses between the two methods:
\begin{itemize}
    \item \textbf{Lineage-Regularized MF}: Robust to incomplete (missing) lineage data (removal of 40\% of links led to only a 10\% decrease in correlation), but highly sensitive to incorrect lineage additions (adding 40\% false links caused a 50\% decrease in correlation). For LRMF, accuracy is better preserved by omitting uncertain lineage information rather than risking incorrect additions.
    \item \textbf{Model Lineage Average}: Demonstrated robustness against both random additions and removals of lineage data. This resilience likely stems from random lineage connections averaging out model-specific variations and pulling results toward a global mean.
\end{itemize}

\begin{figure}[ht]
    \centering
    \includegraphics[width=0.80\textwidth]{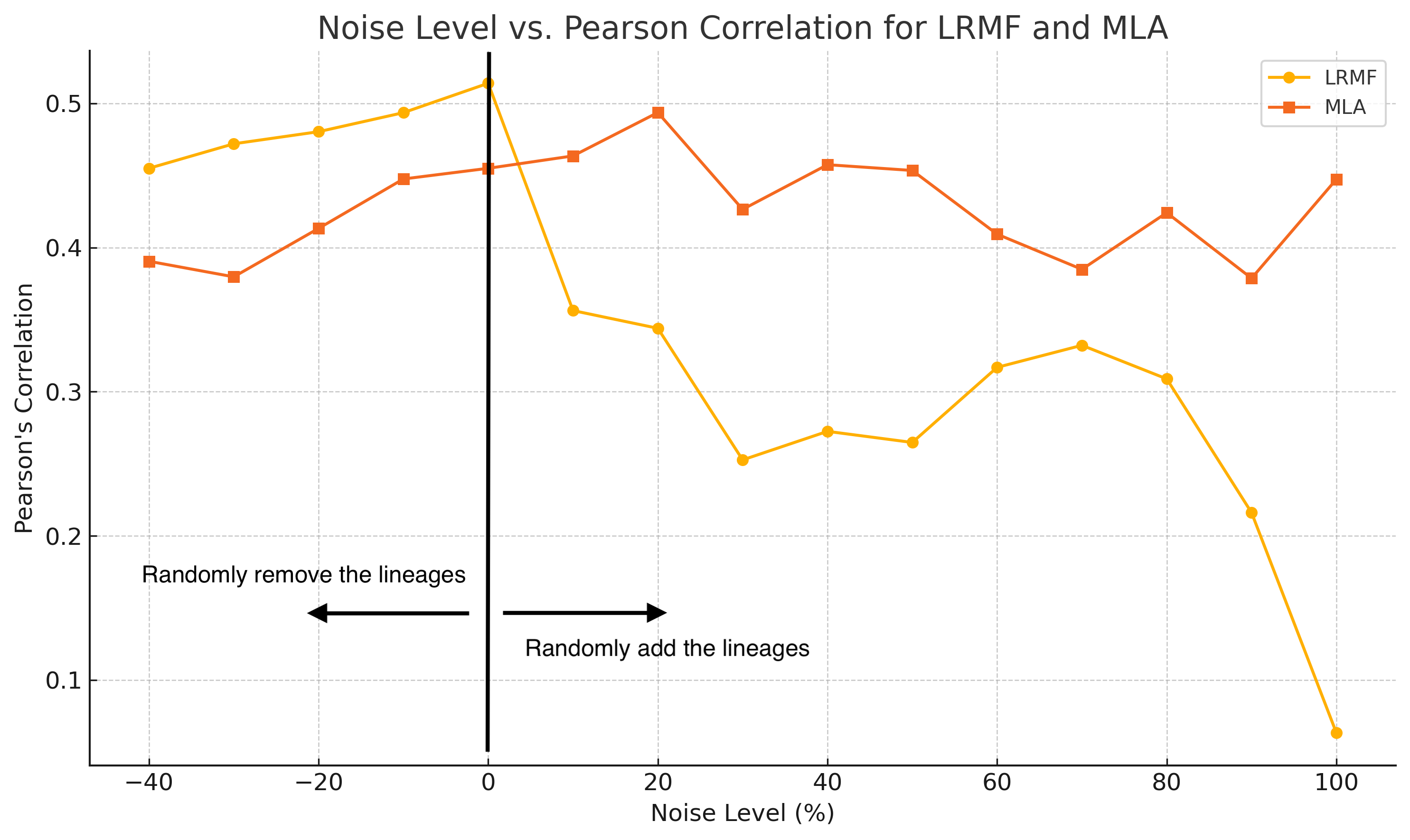}
    \caption{Effect of noise in lineage data on model prediction accuracy in the \emph{heterogeneous} setting, measured by Pearson correlation. Noise is introduced either by randomly removing lineage links (negative values) or randomly adding spurious links (positive values). LRMF shows robustness to missing links but suffers significant degradation when incorrect lineages are added, while MLA remains comparatively stable under both scenarios.}
    \label{fig:lineage_matrix_with_noise}
\end{figure}

\section{Impact of the Number of Observed Instances per Model on Prediction Performance}

We investigate how varying the number of observed instances per model ($t$) affects predictive performance. By sampling $t \in \{5, 10, 20, 50, 100, 200, 500, 1000\}$ instances for each model in the training data, we simulate scenarios ranging from minimal evaluation (resource-constrained) to comprehensive benchmarking.

\begin{figure*}[ht]
  \centering
  \begin{subfigure}[b]{0.49\textwidth}
    \centering
    \includegraphics[width=\textwidth]{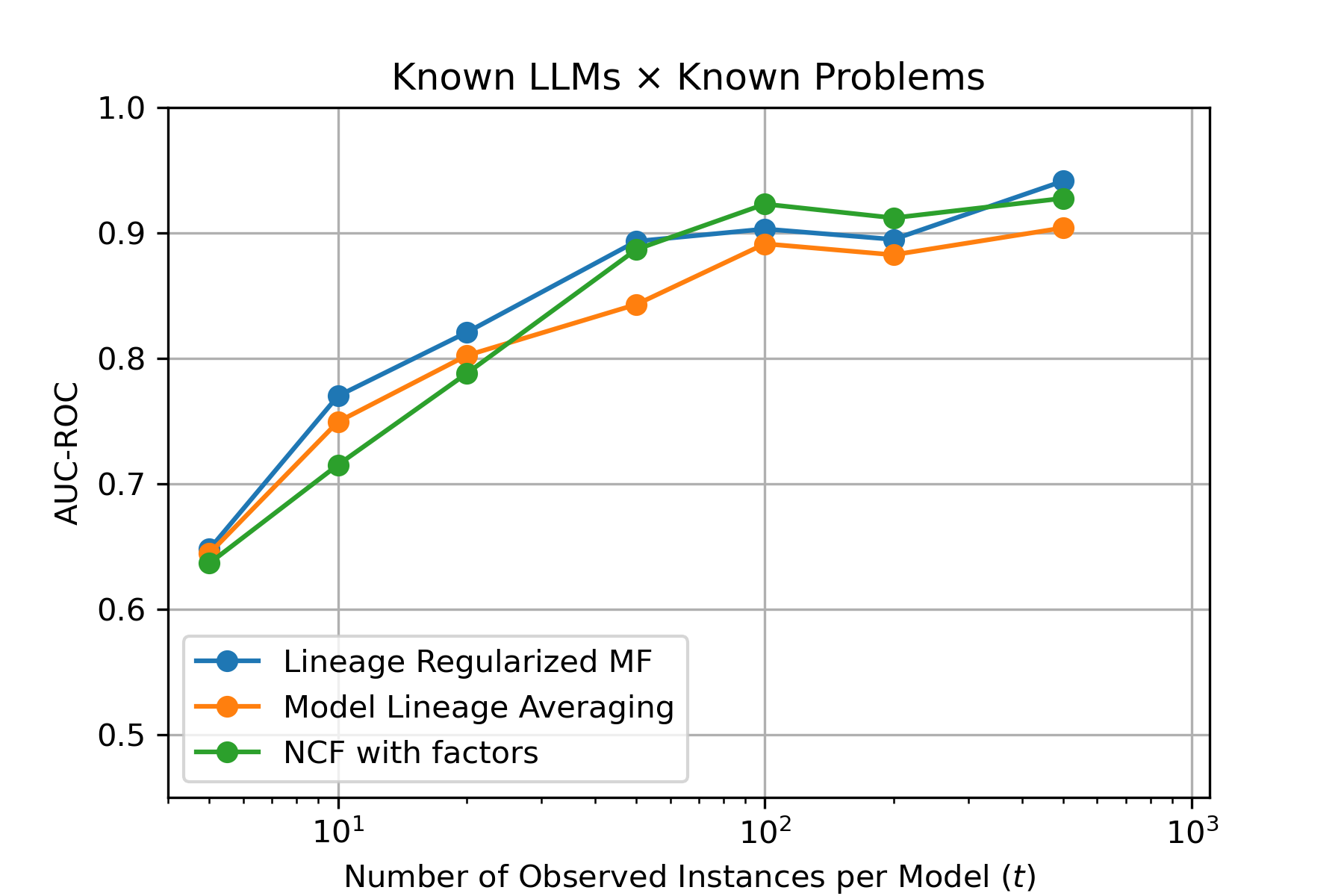}
    \caption{In-distribution prediction: AUC–ROC for withheld instance–model pairs within training models}
    \label{fig:homo_train_auc}
  \end{subfigure}
  \hfill
  \begin{subfigure}[b]{0.49\textwidth}
    \centering
    \includegraphics[width=\textwidth]{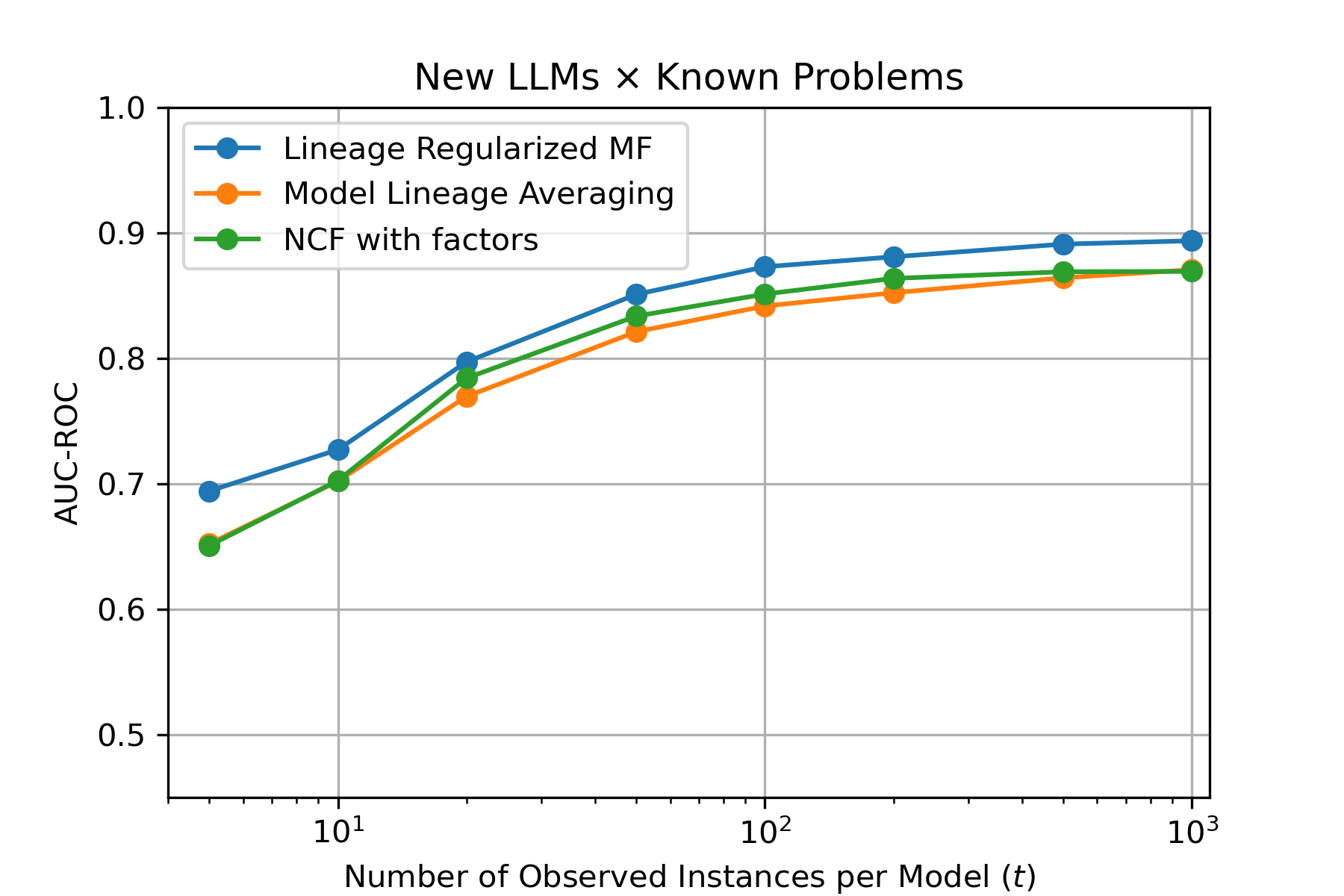}
    \caption{Cold-start prediction: AUC–ROC for instance–model pairs of unseen test models}
    \label{fig:homo_test_auc}
  \end{subfigure}
  \caption{AUC–ROC as a function of the number of observed instances per model ($t$) in the \emph{homogeneous} Qwen 2.5-7B family: (a) in-distribution prediction and (b) cold-start prediction.}
  \label{fig:auc_vs_t_homo}
\end{figure*}

\begin{figure*}[ht]
  \centering
  \begin{subfigure}[b]{0.49\textwidth}
    \centering
    \includegraphics[width=\textwidth]{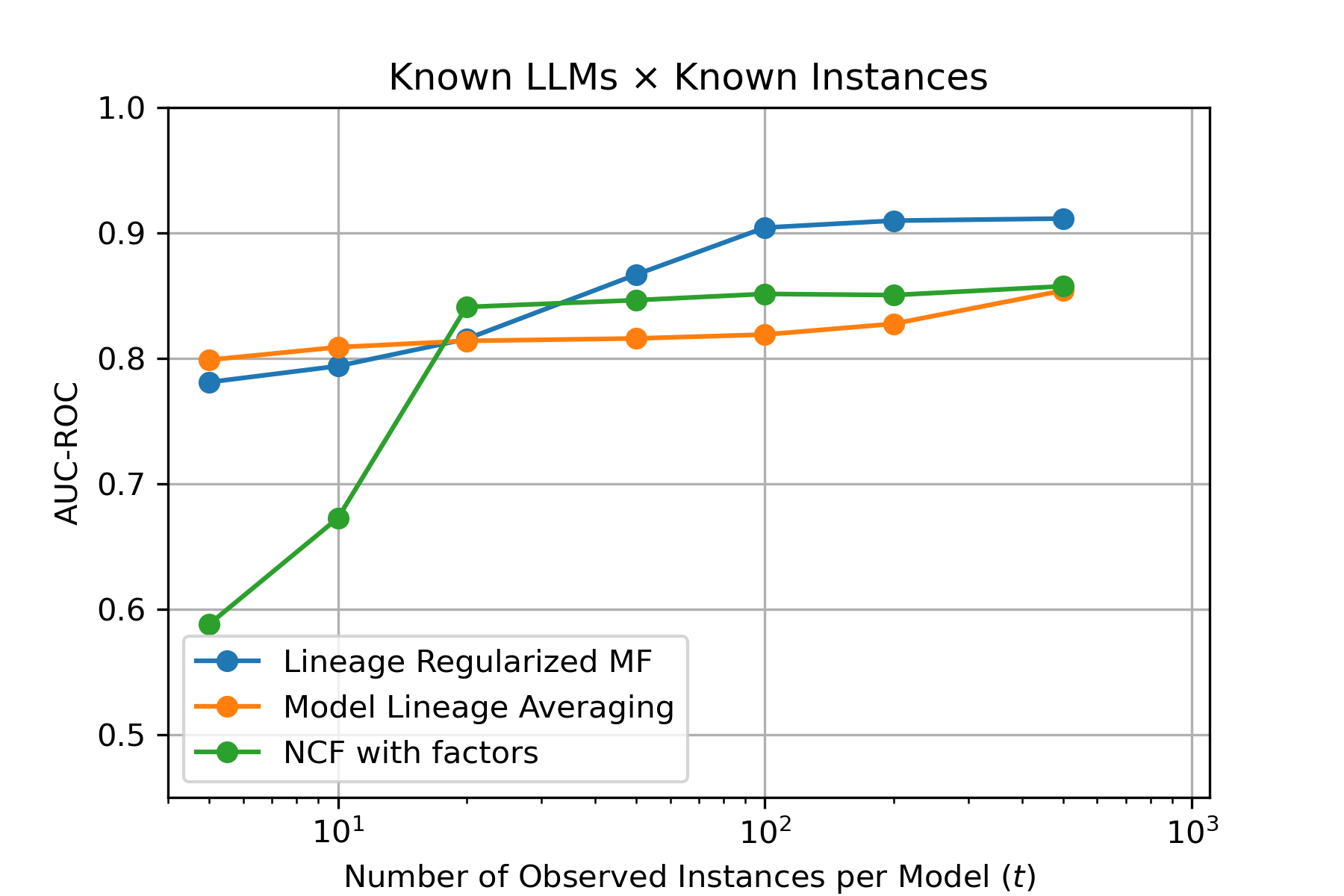}
    \caption{In-distribution prediction: AUC–ROC for withheld instance–model pairs within training models}
    \label{fig:hetero_train_auc}
  \end{subfigure}
  \hfill
  \begin{subfigure}[b]{0.49\textwidth}
    \centering
    \includegraphics[width=\textwidth]{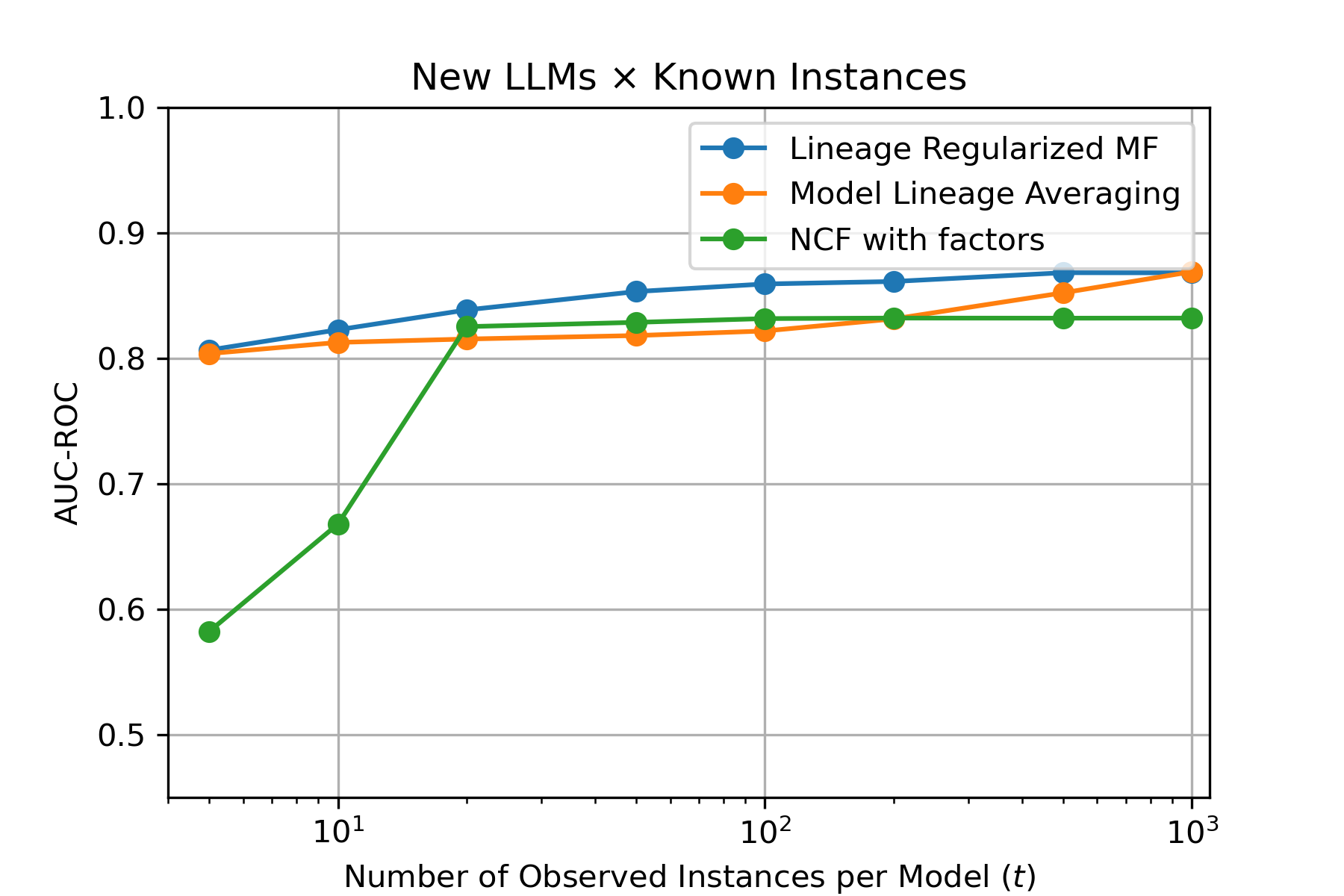}
    \caption{Cold-start prediction: AUC–ROC for instance–model pairs of unseen test models}
    \label{fig:hetero_test_auc}
  \end{subfigure}
  \caption{AUC–ROC as a function of the number of observed instances per model ($t$) in the \emph{heterogeneous} Hugging Face model set: (a) in-distribution prediction and (b) cold-start prediction.}
  \label{fig:auc_vs_t_hetero}
\end{figure*}

Figures~\ref{fig:auc_vs_t_homo} and~\ref{fig:auc_vs_t_hetero} summarize the test-set AUC-ROC across two critical evaluation scenarios for both the \emph{homogeneous} (Qwen 2.5-7B family) and \emph{heterogeneous} (all models) settings:
\begin{itemize}
\item \textbf{In-distribution prediction}: Estimating performance on unobserved instance-model pairs, where the models themselves are included in the training set but certain instances remain unobserved (shown in Figures~\ref{fig:homo_train_auc} and~\ref{fig:hetero_train_auc} for the homogeneous and heterogeneous settings, respectively). While this differs from the main cold-start scenario, it requires imputing missing data due to limited observations per model.
\item \textbf{Cold-start prediction}: Zero-shot performance on entirely new models unseen during training, representing the main cold-start scenario of this work (shown in Figures~\ref{fig:homo_test_auc} and~\ref{fig:hetero_test_auc} for the homogeneous and heterogeneous settings, respectively).
\end{itemize}

The results align with our main findings in Section~\ref{sec:experiments}: Lineage-Regularized MF consistently outperforms baselines, with its advantage becoming more pronounced in the cold-start scenario and with fewer observed instances per model. This confirms our hypothesis that lineage relationships provide valuable signals for predicting the performance of new or sparsely evaluated models.

\begin{figure*}[ht]
  \centering
  \begin{subfigure}[b]{0.49\textwidth}
    \centering
    \includegraphics[width=\textwidth]{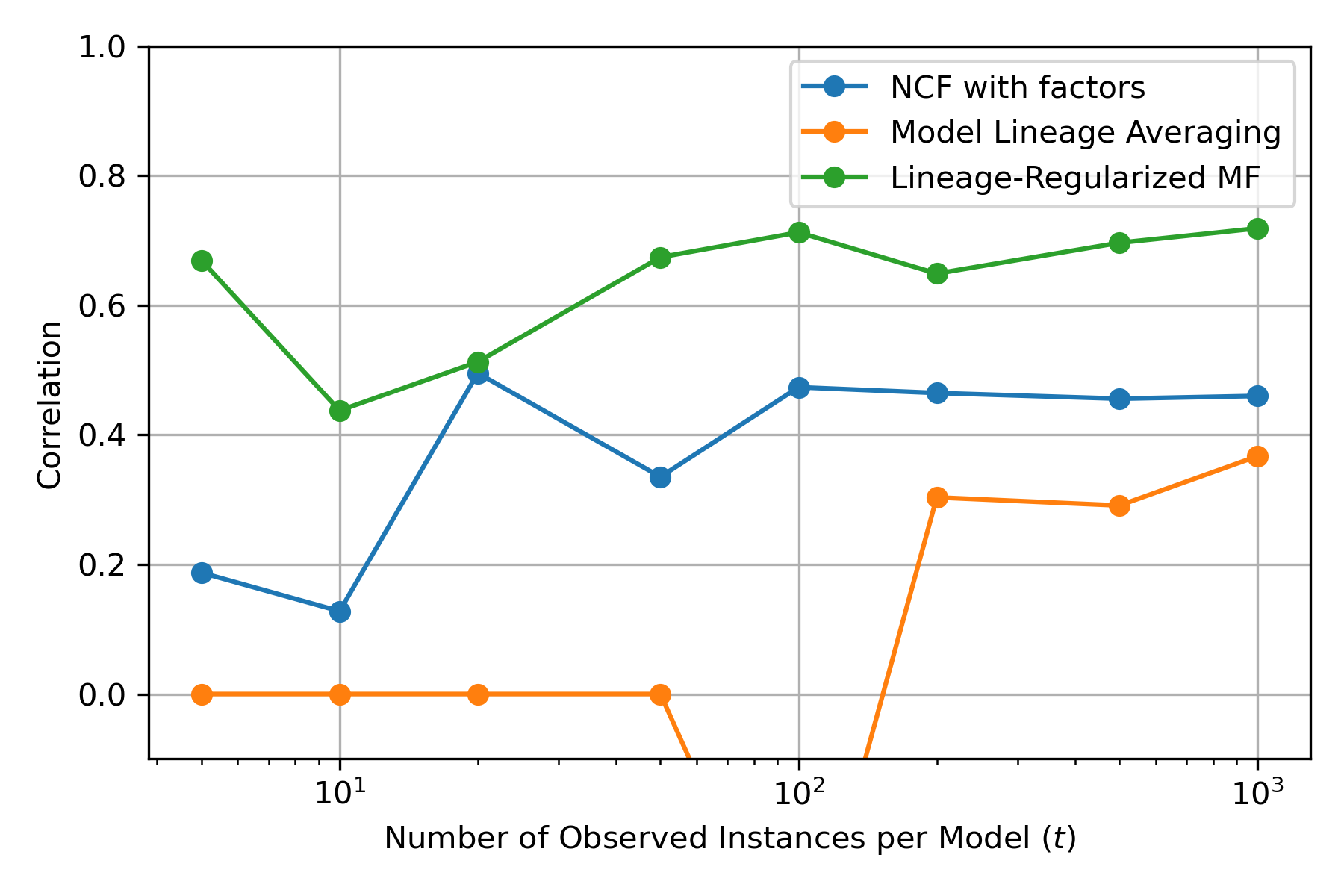}
    \caption{\emph{Homogeneous} setting (Qwen 2.5-7B family): Pearson correlation for cold-start prediction}
    \label{fig:homo_cold_corr}
  \end{subfigure}
  \hfill
  \begin{subfigure}[b]{0.49\textwidth}
    \centering
    \includegraphics[width=\textwidth]{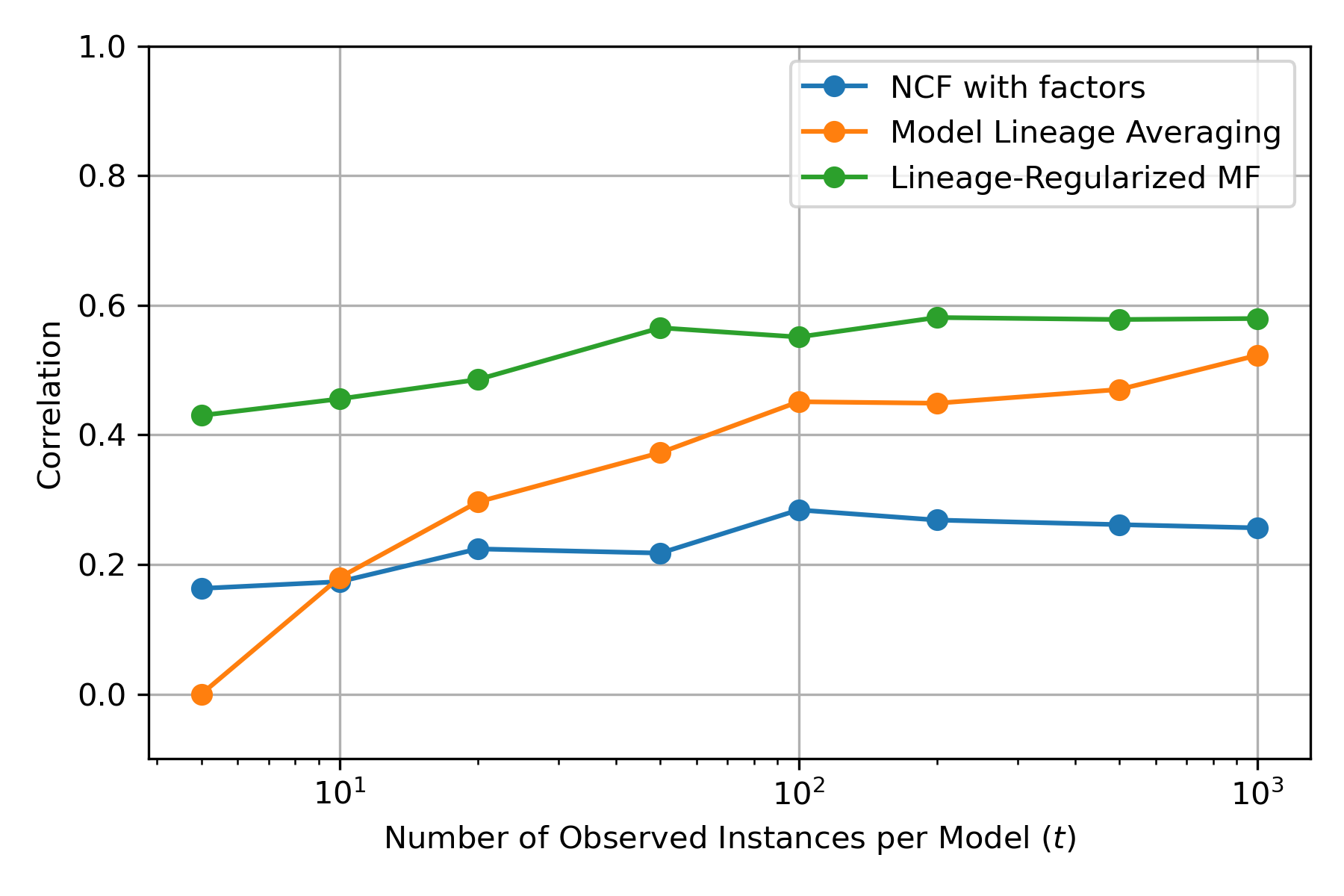}
    \caption{\emph{Heterogeneous} setting (all models): Pearson correlation for cold-start prediction}
    \label{fig:hetero_cold_corr}
  \end{subfigure}
  \caption{Pearson correlation coefficient for cold-start prediction as a function of the number of observed instances per model ($t$): (a) \emph{homogeneous} Qwen 2.5-7B family and (b) \emph{heterogeneous} Hugging Face model set.}
  \label{fig:coldstart_corr_vs_t}
\end{figure*}

\paragraph{Impact of Limited Observations} To quantify the robustness of our approach under data-constrained scenarios, we systematically reduced the number of observed instances per model from 1,000 down to as few as 5 instances. As shown in Figure~\ref{fig:coldstart_corr_vs_t}, Lineage-Regularized MF maintains impressive prediction accuracy even with drastically reduced observations, demonstrating significant resilience compared to Model Lineage Averaging and NCF with factors.

The performance gap between methods is especially pronounced in the homogeneous setting (Figure~\ref{fig:homo_cold_corr}), where Lineage-Regularized MF achieves with just 50-100 instances per model what baseline methods require 500+ instances to match. This 5-10x reduction in required evaluations translates to substantial computational savings during model development. In the heterogeneous setting (Figure~\ref{fig:hetero_cold_corr}), while all methods benefit from the diversity of models, our approach still maintains a consistent advantage across all observation levels.

These results have significant practical implications: developers can reliably predict a new model's performance across multiple benchmarks after evaluating it on just 50-100 carefully selected instances, provided that lineage information is properly incorporated. This dramatically reduces the computational overhead typically associated with comprehensive benchmarking during iterative LLM development.

\section{Routing Details}
\begin{figure}[ht]
    \centering
    \includegraphics[width=0.90\textwidth]{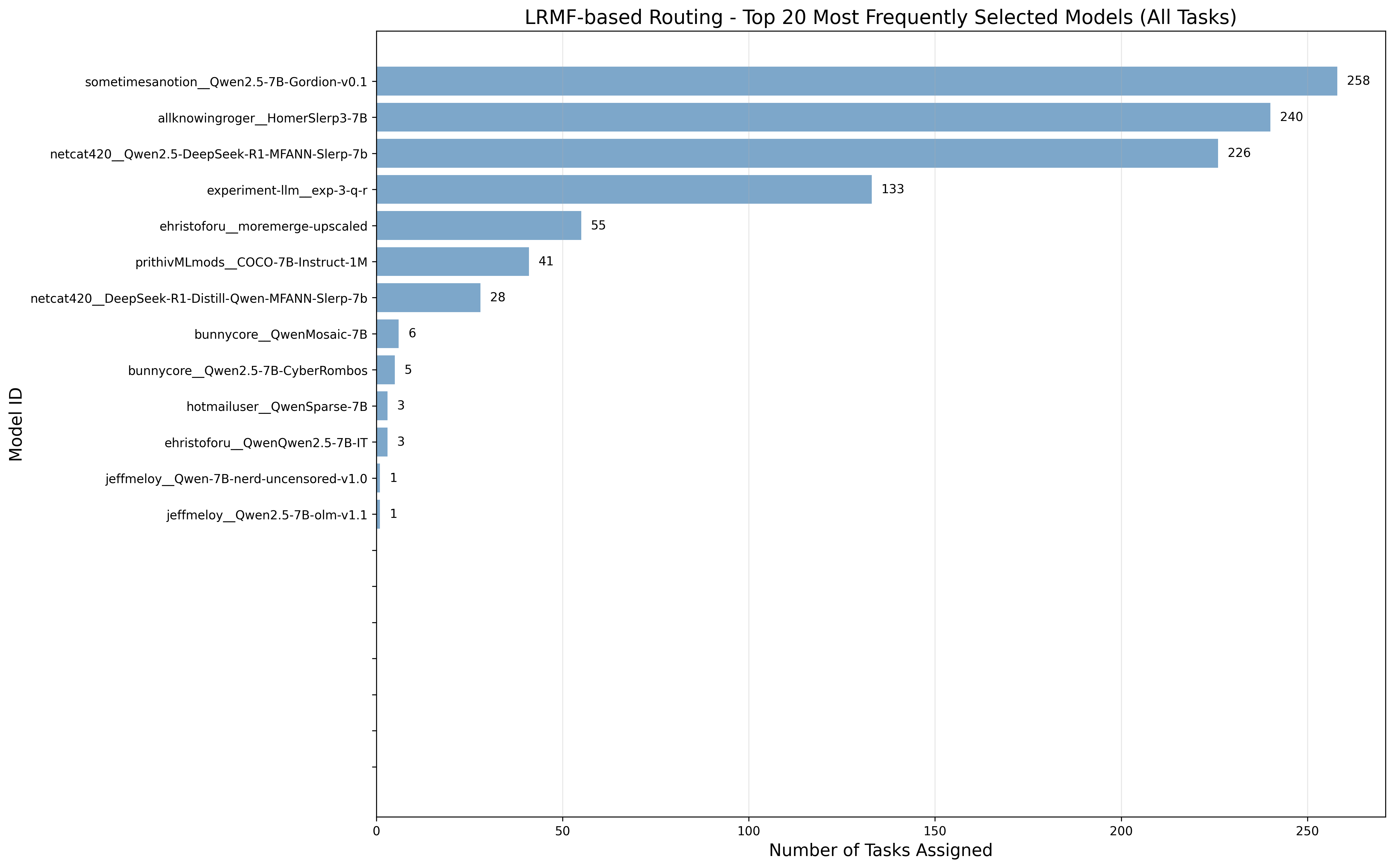}
    \caption{Top 20 most frequently routed-to models by LRMF-based routing in the \emph{homogeneous} experiment.}
    \label{fig:qwen_model_assignment_all_tasks_graphrmf}
\end{figure}

\begin{figure}[ht]
    \centering
    \includegraphics[width=0.90\textwidth]{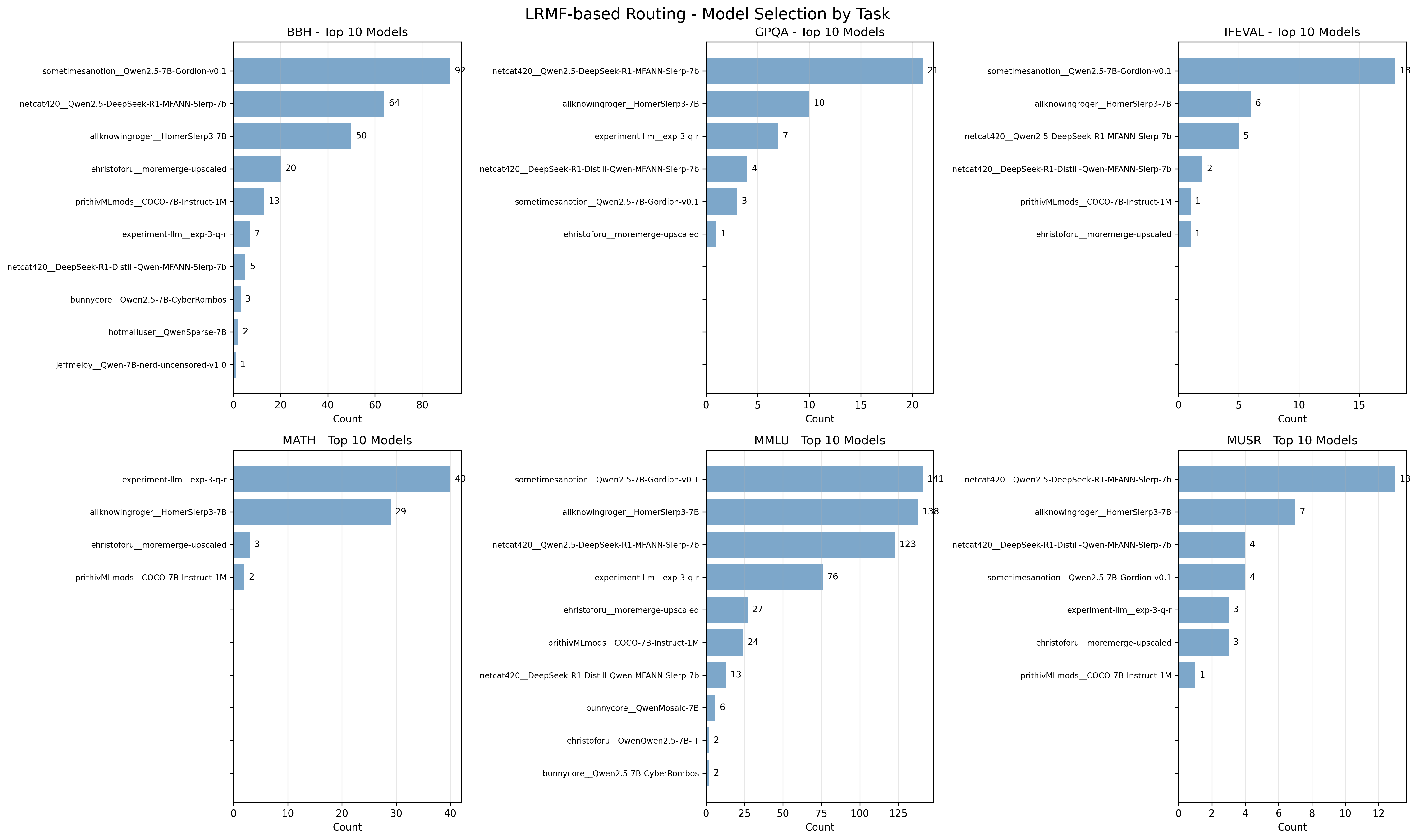}
    \caption{Top 10 most frequently routed-to models by LRMF-based routing in the \emph{homogeneous} experiment. This demonstrates that this routing approach assigns different models as optimal destinations depending on the benchmark.}
    \label{fig:qwen_model_assignment_by_task_graphrmf}
\end{figure}

\begin{figure}[ht]
    \centering
    \includegraphics[width=0.90\textwidth]{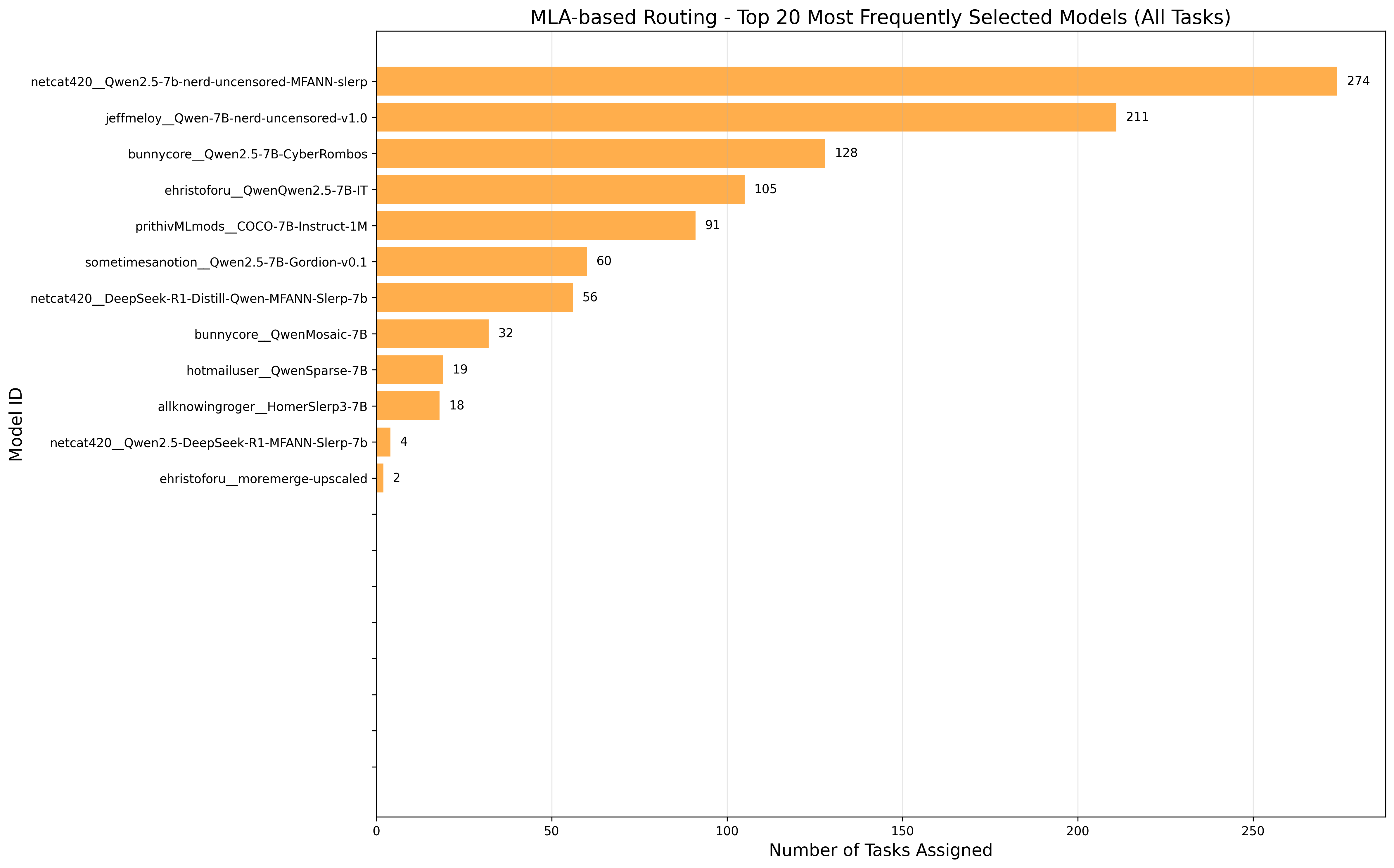}
    \caption{Top 20 most frequently routed-to models by MLA-based routing in the \emph{homogeneous} experiment.}
    \label{fig:qwen_model_assignment_all_tasks_mknnrouting}
\end{figure}

\begin{figure}[ht]
    \centering
    \includegraphics[width=0.90\textwidth]{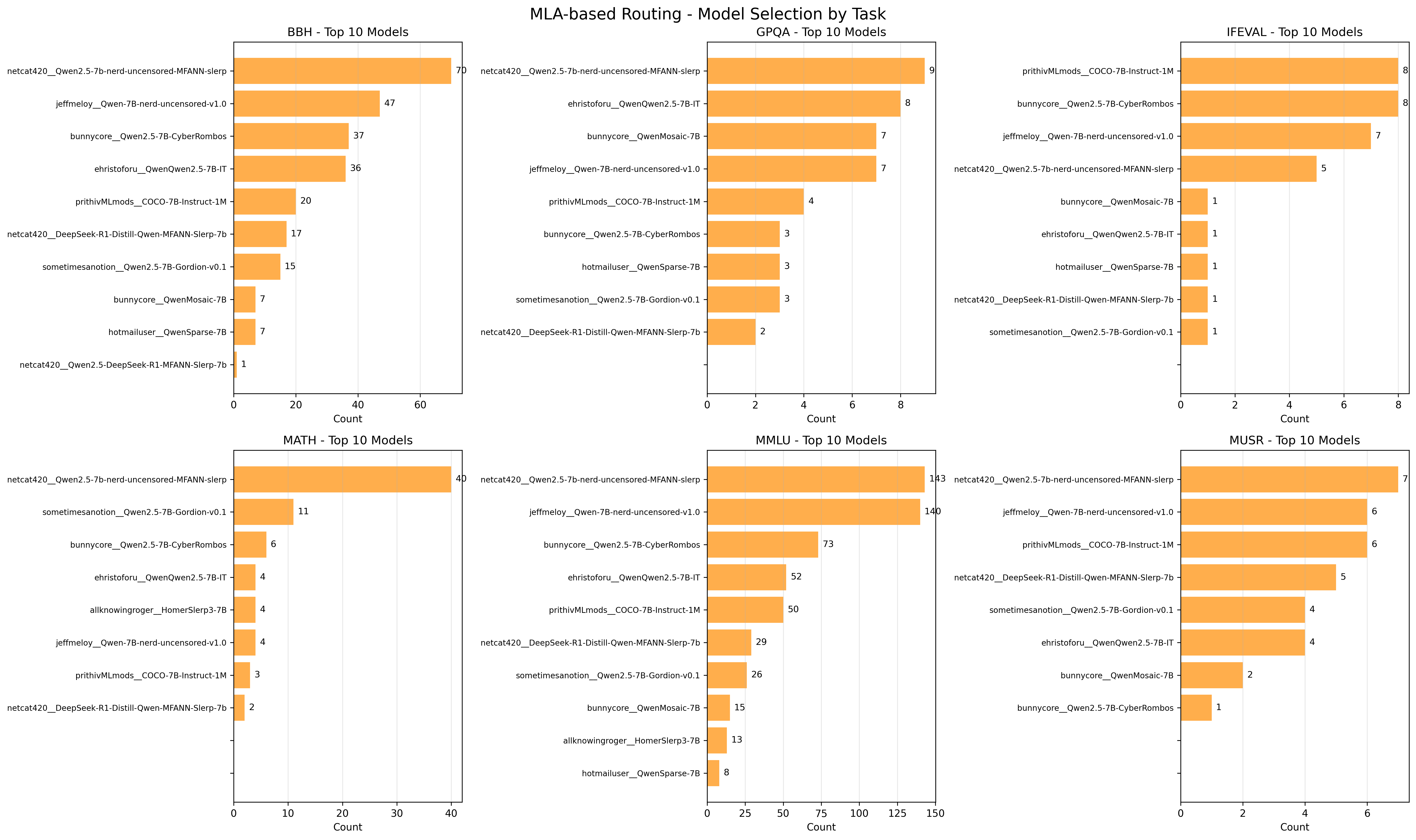}
    \caption{Top 10 most frequently routed-to models by MLA-based routing in the \emph{homogeneous} experiment. This demonstrates that this routing approach assigns different models as optimal destinations depending on the benchmark.}
    \label{fig:qwen_model_assignment_by_task_mknnrouting}
\end{figure}

\begin{figure}[ht]
    \centering
    \includegraphics[width=0.90\textwidth]{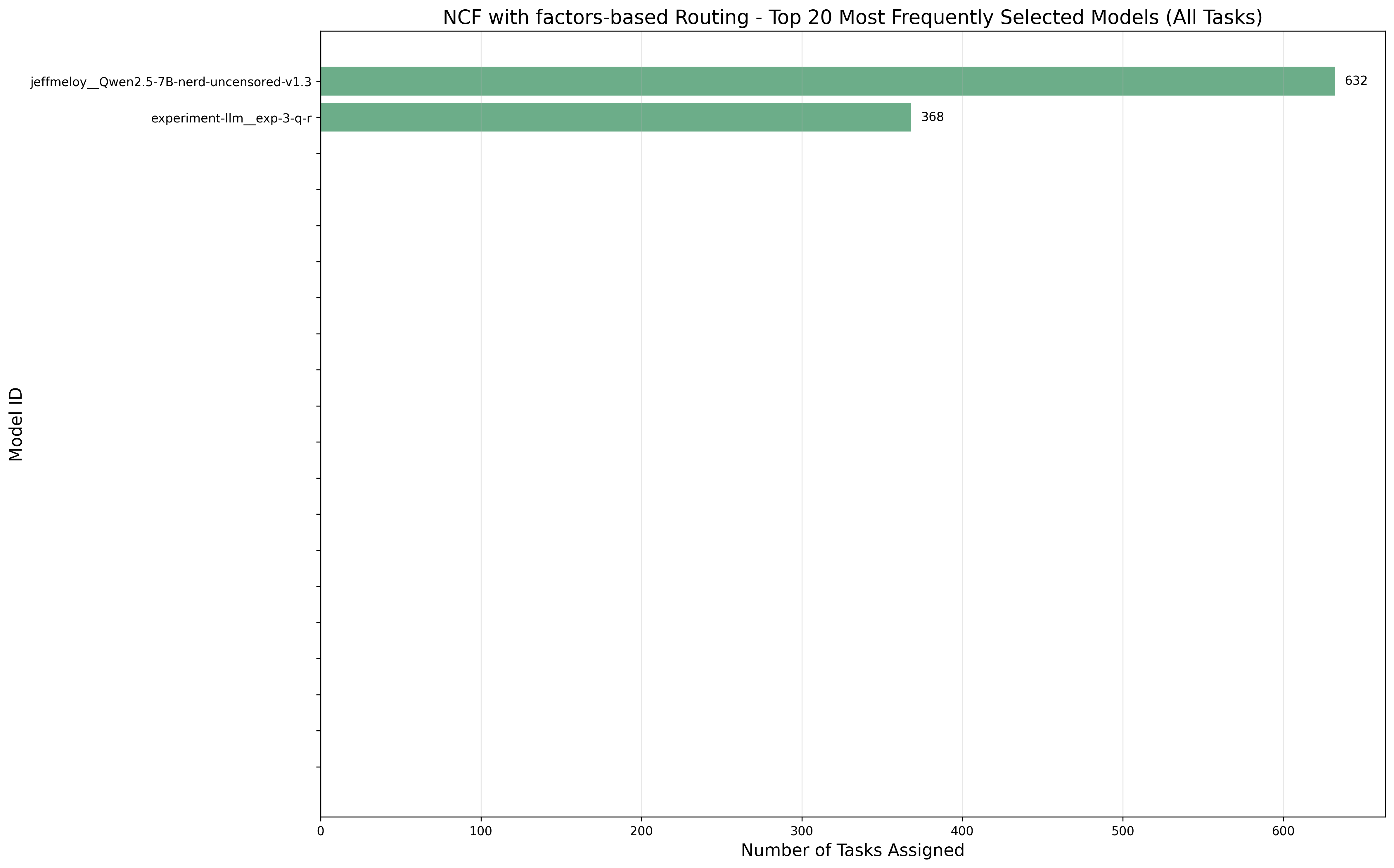}
    \caption{Top 20 most frequently routed-to models by NCF with factors-based routing in the \emph{homogeneous} experiment. This result indicates that the routing strategy assigns all tasks exclusively to two models.}
    \label{fig:qwen_model_assignment_all_tasks_ncf_f}
\end{figure}

\begin{figure}[ht]
    \centering
    \includegraphics[width=0.90\textwidth]{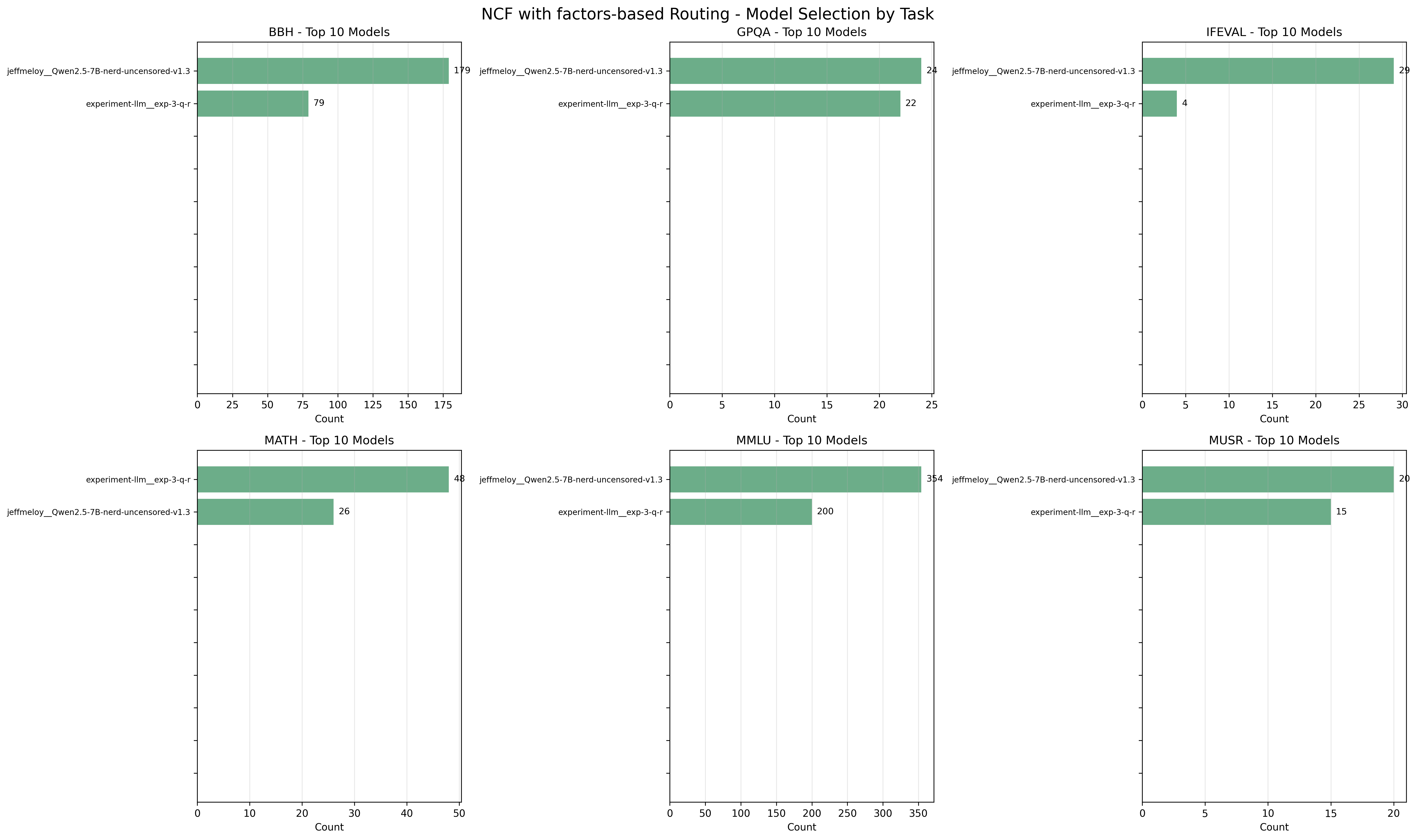}
    \caption{Top 10 most frequently routed-to models by NCF with factors-based routing in the \emph{homogeneous} experiment. This result indicates that the routing strategy assigns all tasks exclusively to two models.}
    \label{fig:qwen_model_assignment_by_task_ncf_f}
\end{figure}

\begin{figure}[ht]
    \centering
    \includegraphics[width=0.90\textwidth]{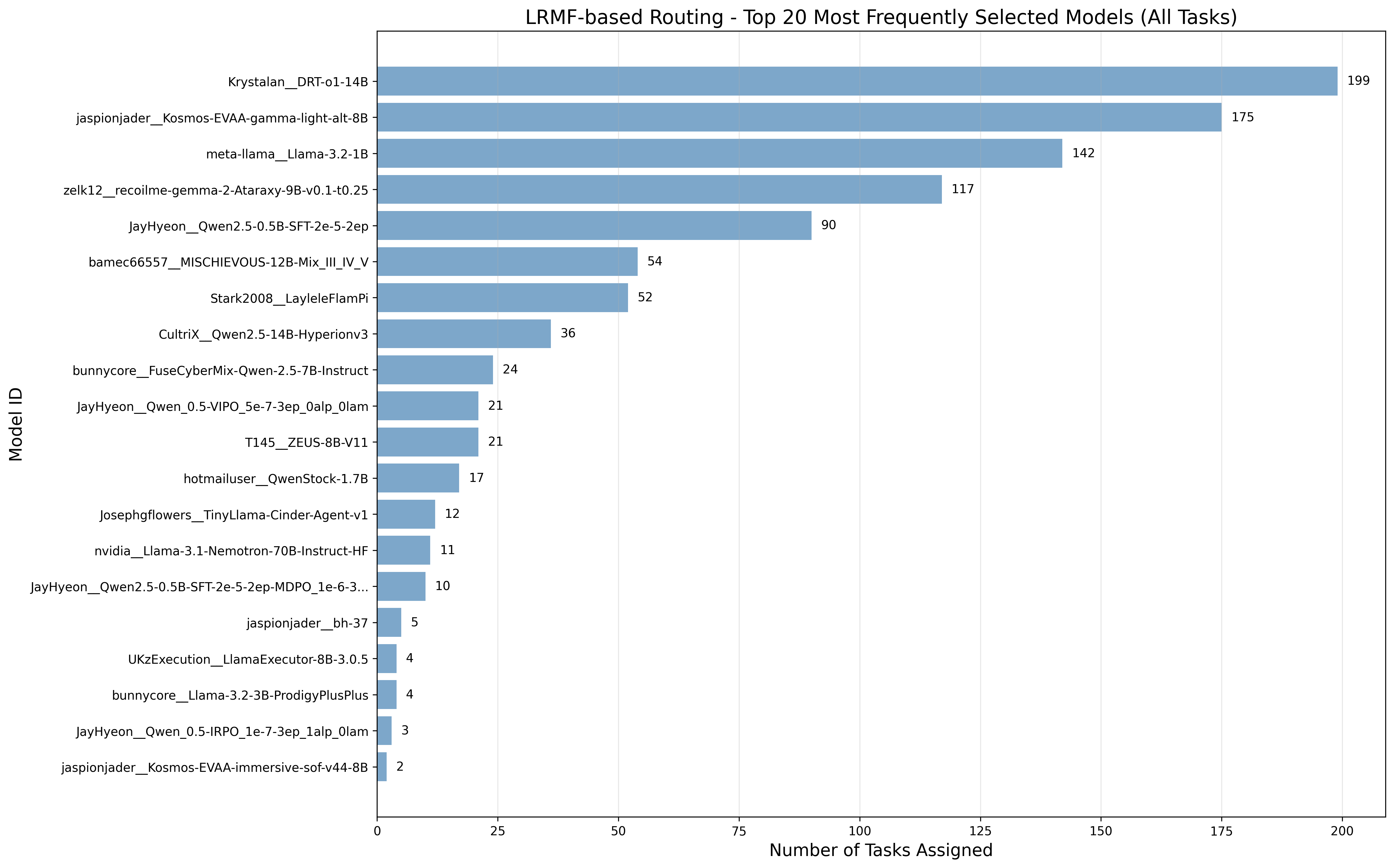}
    \caption{Top 20 most frequently routed-to models by LRMF-based routing in the \emph{heterogeneous} experiment.}
    \label{fig:all_model_assignment_all_tasks_graphrmf}
\end{figure}

\begin{figure}[ht]
    \centering
    \includegraphics[width=0.90\textwidth]{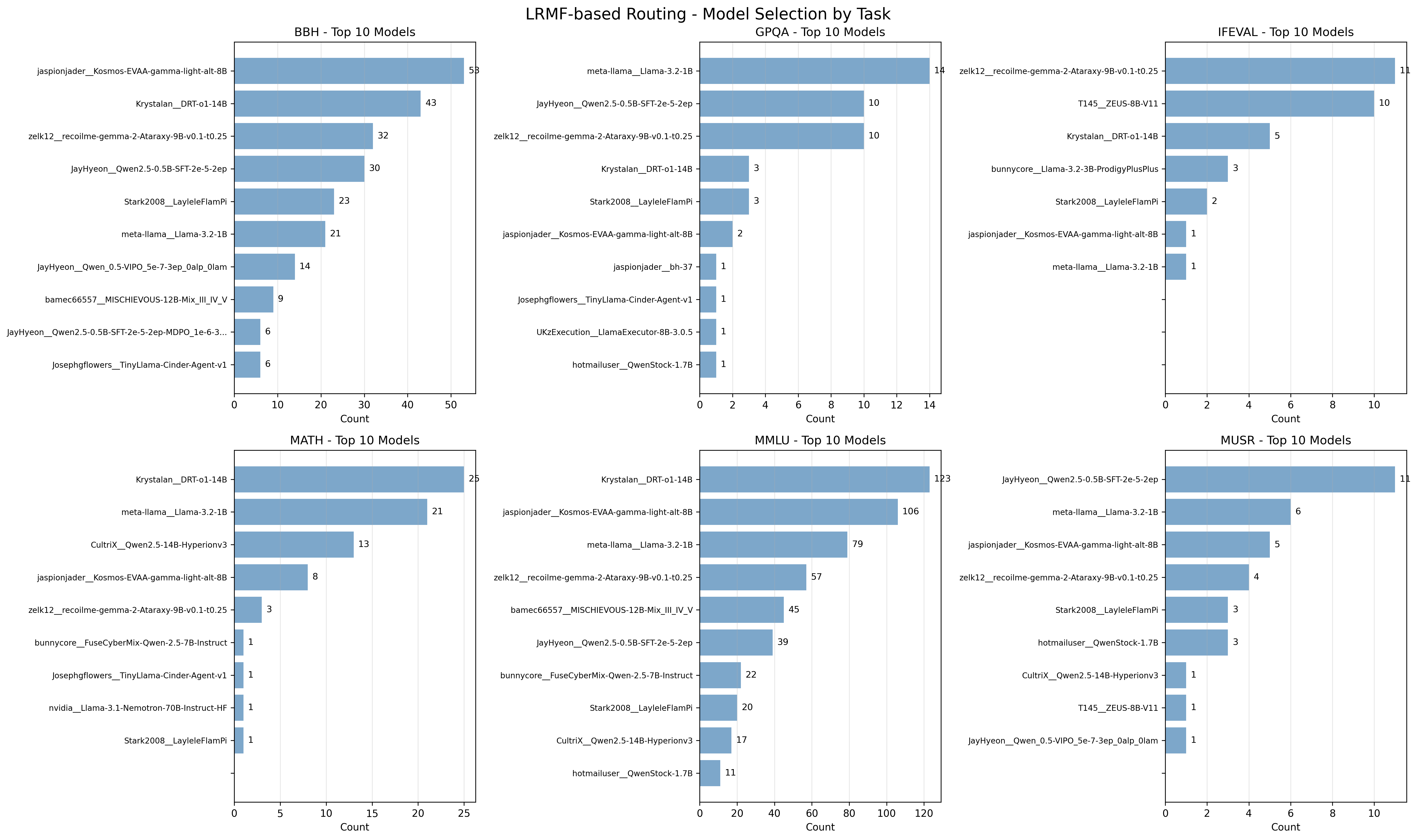}
    \caption{Top 10 most frequently routed-to models by LRMF-based routing in the \emph{heterogeneous} experiment. This demonstrates that this routing approach assigns different models as optimal destinations depending on the benchmark.}
    \label{fig:all_model_assignment_by_task_graphrmf}
\end{figure}

\begin{figure}[ht]
    \centering
    \includegraphics[width=0.90\textwidth]{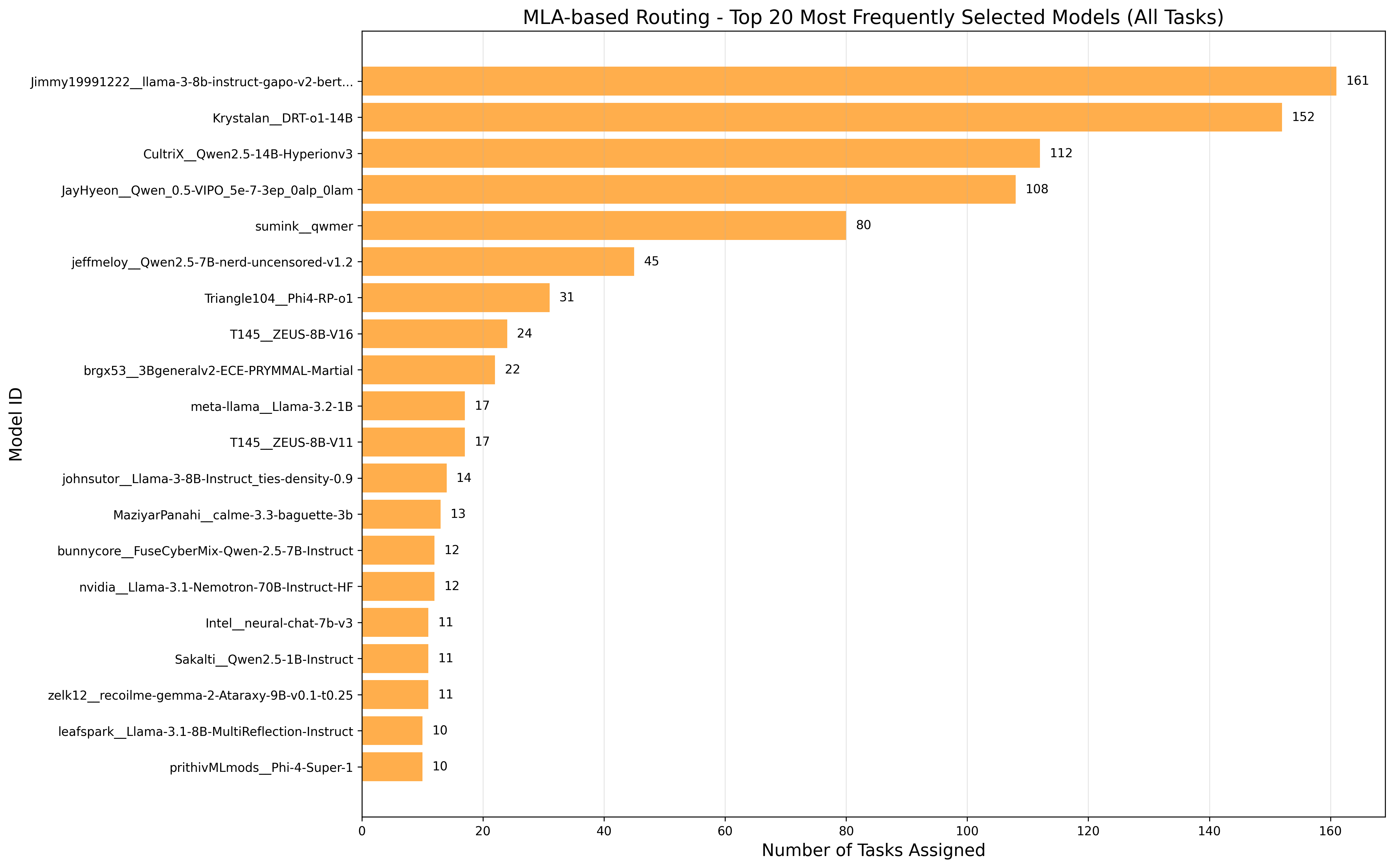}
    \caption{Top 20 most frequently routed-to models by MLA-based routing in the \emph{heterogeneous} experiment.}
    \label{fig:all_model_assignment_all_tasks_mknnrouting}
\end{figure}

\begin{figure}[ht]
    \centering
    \includegraphics[width=0.90\textwidth]{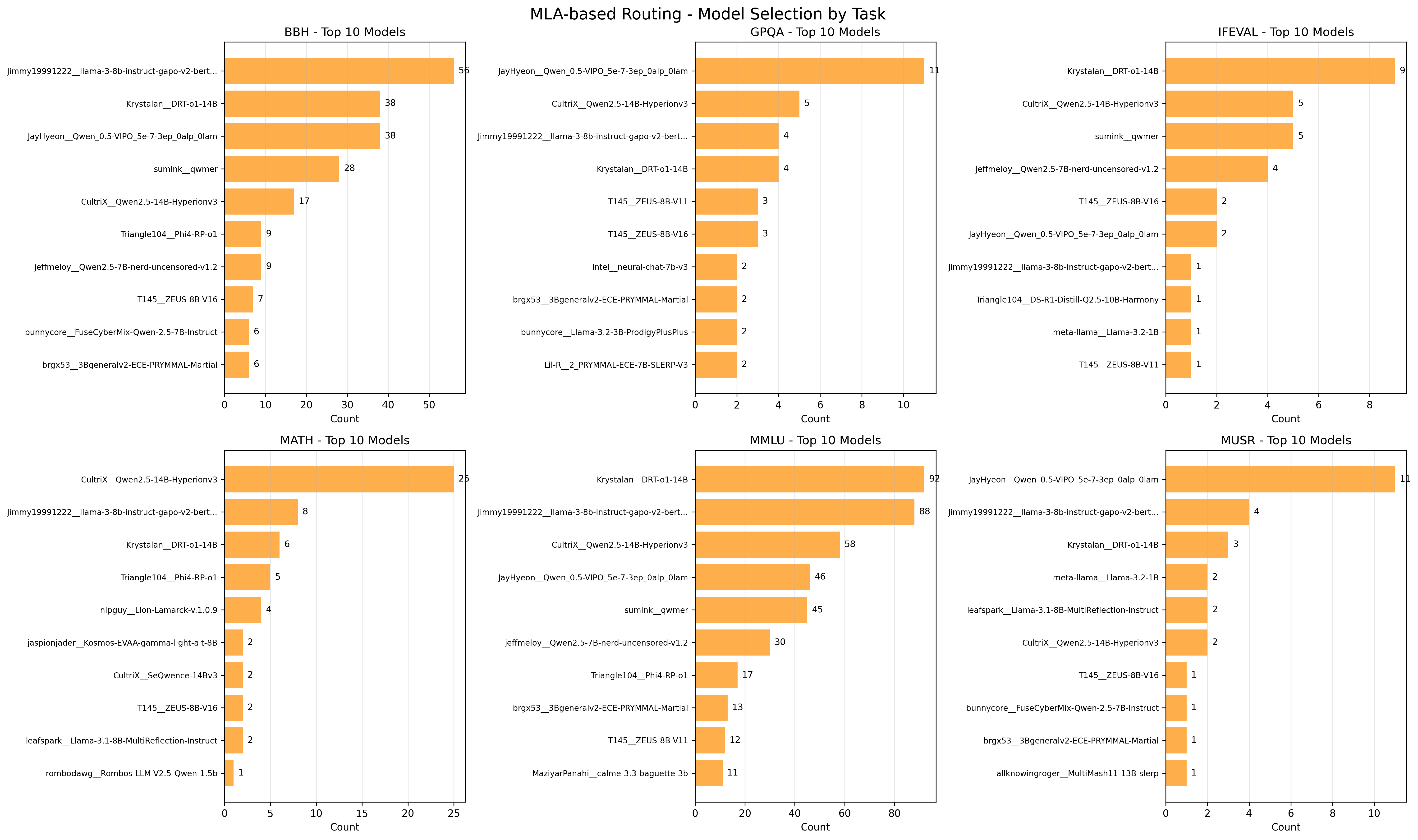}
    \caption{Top 10 most frequently routed-to models by MLA-based routing in the \emph{heterogeneous} experiment. This demonstrates that this routing approach assigns different models as optimal destinations depending on the benchmark.}
    \label{fig:all_model_assignment_by_task_mknnrouting}
\end{figure}

\begin{figure}[ht]
    \centering
    \includegraphics[width=0.90\textwidth]{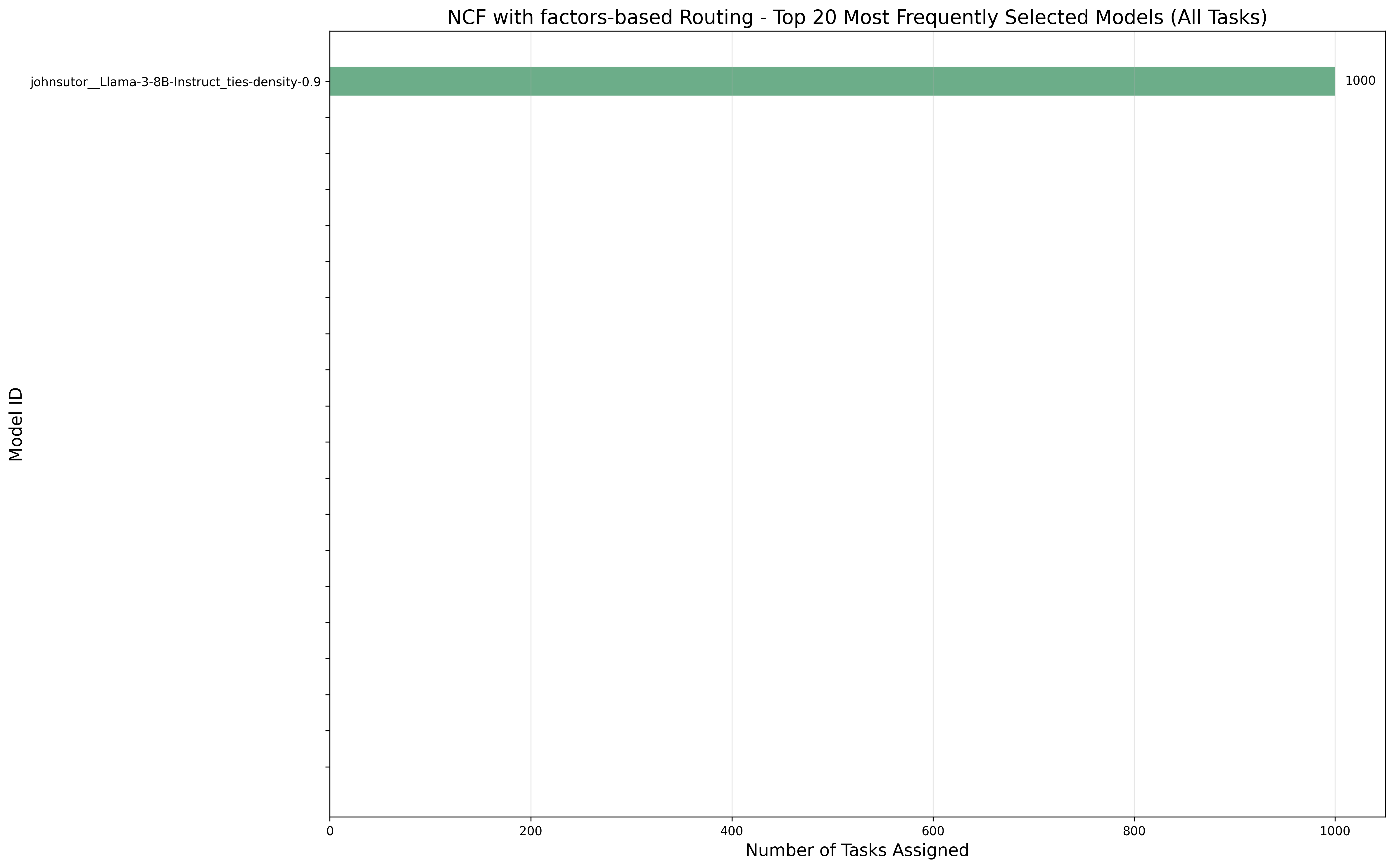}
    \caption{Top 20 most frequently routed-to models by NCF with factors-based routing in the \emph{heterogeneous} experiment. This result indicates that the routing strategy assigns all tasks exclusively to one model.}
    \label{fig:all_model_assignment_all_tasks_ncf_f}
\end{figure}

\begin{figure}[ht]
    \centering
    \includegraphics[width=0.90\textwidth]{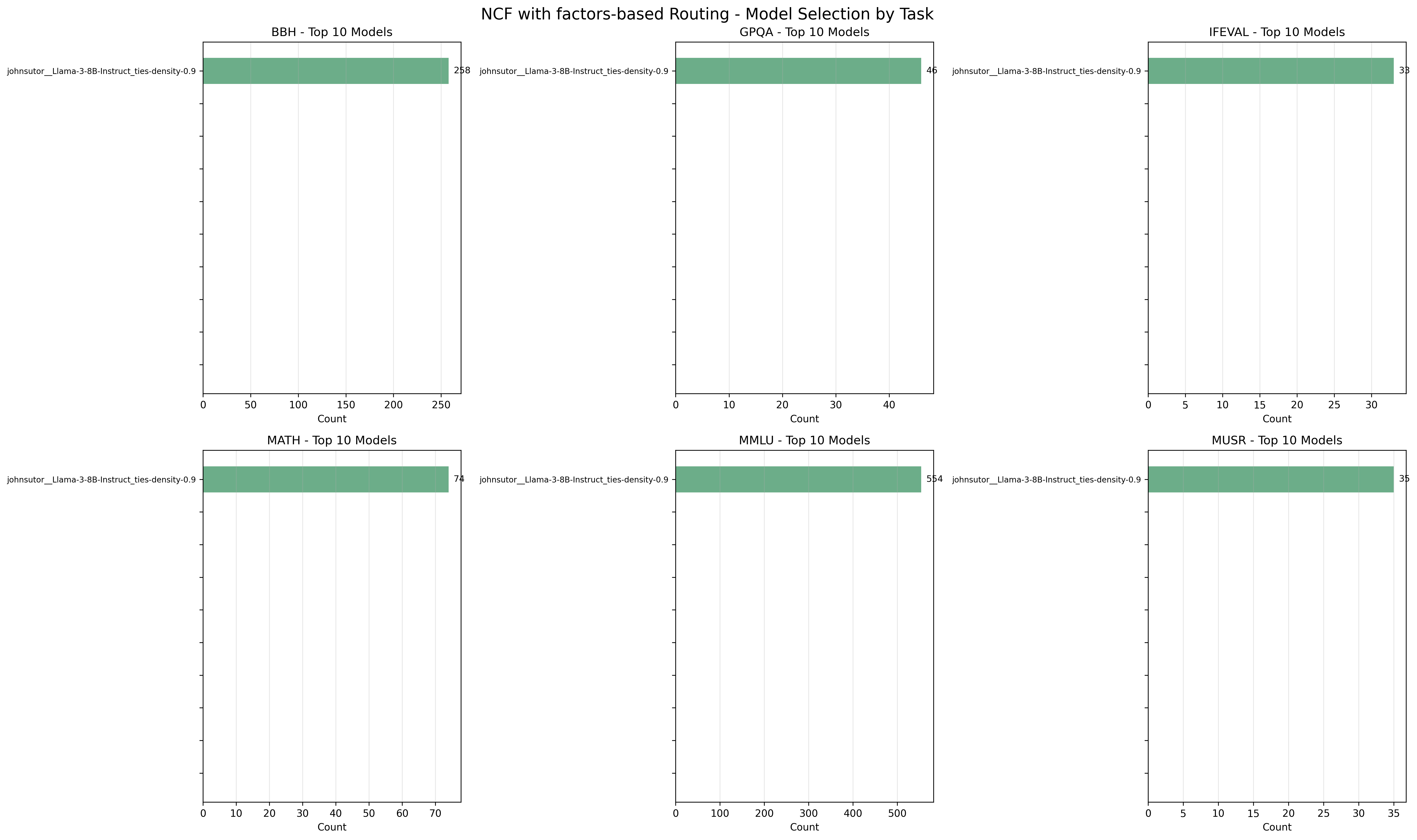}
    \caption{Top 10 most frequently routed-to models by NCF with factors-based routing in the \emph{heterogeneous} experiment. This result indicates that the routing strategy assigns all tasks exclusively to one model.}
    \label{fig:all_model_assignment_by_task_ncf_f}
\end{figure}

Below we provide a detailed analysis of the routing assignments produced by each method. For both the heterogeneous model set (all models) and the homogeneous Qwen-2.5-7B family, we report:
\begin{enumerate}
  \item The top-20 models most frequently selected across \emph{all} test instances (Figures \ref{fig:all_model_assignment_all_tasks_graphrmf}–\ref{fig:all_model_assignment_all_tasks_ncf_f}, Figures \ref{fig:qwen_model_assignment_all_tasks_graphrmf}–\ref{fig:qwen_model_assignment_all_tasks_ncf_f}).
  \item The top-20 models most frequently selected \emph{per benchmark} (Figures \ref{fig:all_model_assignment_by_task_graphrmf}–\ref{fig:all_model_assignment_by_task_ncf_f}, Figures \ref{fig:qwen_model_assignment_by_task_graphrmf}–\ref{fig:qwen_model_assignment_by_task_ncf_f}).
\end{enumerate}

In both the homogeneous (Qwen-2.5-7B family) and heterogeneous (all models) settings, we observe the following common routing behaviors:

\paragraph{LRMF-Based Routing.}
LRMF consistently leverages lineage information to select a diverse set of high-performing variants. When aggregated across all tasks, instruction-tuned and merge-derived (“soup”) models dominate the top selections. On a per-benchmark basis, the method dynamically adapts—for example, favoring instruction-tuned models on IFEval and large base models on reasoning benchmarks such as BBH and MMLU-Pro.

\paragraph{Model Lineage Averaging–Based Routing.}
The kNN-style lineage averaging approach concentrates almost exclusively on immediate ancestor checkpoints. It shows minimal variability across benchmarks, typically defaulting to whichever parent model achieved the highest raw score.

\paragraph{NCF with Factors–Based Routing.}
The factor-based MLP exhibits an overwhelming bias toward a single model across both settings. This behavior indicates overfitting to dominant metadata signals and a lack of task-specific discrimination.

\subsection{\emph{Homogeneous} Qwen-2.5-7B Family setting}
\subsubsection{LRMF-Based Routing}

\clearpage
\subsubsection{Model Lineage Averaging–Based Routing}

\clearpage
\subsubsection{NCF with Factors–Based Routing}

\clearpage
\subsection{\emph{Heterogeneous} Models setting}
\subsubsection{LRMF-Based Routing}

\clearpage
\subsubsection{Model Lineage Averaging–Based Routing}

\clearpage
\subsubsection{NCF with Factors–Based Routing}

\end{document}